%% file: main.tex
\pdfoutput=1
\PassOptionsToPackage{table,dvipsnames}{xcolor}

\documentclass[11pt]{article}

\usepackage[final]{acl}

\usepackage[table,dvipsnames]{xcolor}
\usepackage{collcell}

\usepackage{times}
\usepackage{latexsym}

\usepackage[utf8]{inputenc} 
\usepackage[T1]{fontenc}

\usepackage{paralist}
\usepackage{microtype}

\usepackage{inconsolata}

\usepackage{times,latexsym}
\usepackage{url}
\usepackage{pgf}
\usepackage{booktabs}
\usepackage{enumitem}
\usepackage{expex}
\usepackage[linguistics]{forest}

\usepackage{extdash} 

\definecolor{Gray}{gray}{0.9}
\definecolor{cb-blue-green} {RGB}{ 0,  073,  073}
\definecolor{cb-green-sea}  {RGB}{ 0, 146, 146}
\definecolor{cb-rose}       {RGB}{255, 109, 182}
\definecolor{cb-salmon-pink}{RGB}{255, 182, 119}
\definecolor{cb-purple}     {RGB}{ 73,   0, 146}
\definecolor{cb-blue}       {RGB}{ 0, 109, 219}
\definecolor{cb-lilac}      {RGB}{182, 109, 255}
\definecolor{cb-blue-sky}   {RGB}{109, 182, 255}
\definecolor{cb-blue-light} {RGB}{182, 219, 255}
\definecolor{cb-burgundy}   {RGB}{146,   0,   0}
\definecolor{cb-brown}      {RGB}{146,  73,   0}
\definecolor{cb-clay}       {RGB}{219, 209,   0}
\definecolor{cb-green-lime} {RGB}{ 36, 255,  36}
\definecolor{cb-yellow}     {RGB}{255, 255, 109}
\definecolor{cb-grey}       {RGB}{233, 233, 233}

\usepackage{amsmath,amssymb, amsthm}

\usepackage{graphicx,scalerel} 
\graphicspath{{./images/}} 
\usepackage{subfig}
\usepackage{adjustbox}

\usepackage{tikz}
\usepackage{amssymb}
\usepackage{pifont}
\newcommand{\cmark}{\textcolor{cb-green-sea}{\ding{51}}}%
\newcommand{\xmark}{\textcolor{cb-burgundy}{\ding{55}}}%

\usepackage{booktabs}
\usepackage{multicol}
\usepackage{multirow}
\usepackage{xspace}
\usepackage{arydshln}
\usepackage{makecell}
\usepackage[normalem]{ulem}
\usepackage{array}
\usepackage{tabularx}
\usepackage[most]{tcolorbox}

\usepackage{tcolorbox}
\tcbuselibrary{most}
\definecolor{unchangedcolor}{RGB}{173,216,230}  
\definecolor{changedcolor}{RGB}{255,182,193}  

\newtcbox{\unchangedbox}[1][]{%
    enhanced,
    nobeforeafter,
    tcbox raise base,
    boxrule=0.4pt,
    top=0pt,
    bottom=0pt,
    right=0pt,
    left=0pt,
    arc=3pt,
    boxsep=2pt,
    before upper={\vphantom{A}},
    colback=unchangedcolor,
    #1
}

\newtcbox{\changedbox}[1][]{%
    enhanced,
    nobeforeafter,
    tcbox raise base,
    boxrule=0.4pt,
    top=0pt,
    bottom=0pt,
    right=0pt,
    left=0pt,
    arc=3pt,
    boxsep=2pt,
    before upper={\vphantom{A}},
    colback=changedcolor,
    #1
}

\newtcolorbox{promptbox}[1][]{
    enhanced,
    nobeforeafter,
    boxrule=0.4pt,
    arc=3pt,
    boxsep=2pt,
    left=1pt,
    right=1pt,
    top=1pt,
    bottom=1pt,
    colframe=blue!75!black,
    colback=blue!5,
    #1
}

\newtcolorbox{llmtextbox}[1][]{
    enhanced,
    nobeforeafter,
    boxrule=0.4pt,
    arc=3pt,
    boxsep=2pt,
    left=1pt,
    right=1pt,
    top=1pt,
    bottom=1pt,
    colframe=green!75!black,
    colback=green!5,
    #1
}

\author{Ekaterina Artemova$^1$\hspace{0.2in}
        Jason Lucas$^2$\hspace{0.2in}
        Saranya Venkatraman$^2$ \\
        \textbf{Jooyoung Lee$^2$} \hspace{0.2in}
        \textbf{Sergei Tilga$^1$} \hspace{0.2in}
        \textbf{Adaku Uchendu$^3$} \hspace{0.2in}
        \textbf{Vladislav Mikhailov$^4$} \vspace{0.1in} \\
        $^1$Toloka AI, 
        $^2$The Pennsylvania State University,  
        $^3$MIT Lincoln Laboratory, 
        $^4$University of Oslo
        \vspace{0.1in} \\
    \small{
    \textbf{Correspondence:} \href{mailto:katya-art@toloka.ai}{\texttt{katya-art@toloka.ai}}
}}

\newcommand*{\MinNumber}{30.0}%
\newcommand*{\MidNumber}{60.0} %
\newcommand*{\MaxNumber}{100.0}%

\newcommand{\ApplyGradient}[1]{%
        \ifdim #1 pt > \MidNumber pt
            \pgfmathsetmacro{\PercentColor}{max(min(100.0*(#1 - \MidNumber)/(\MaxNumber-\MidNumber),100.0),0.00)} %
            \hspace{-0.33em}\colorbox{SeaGreen!\PercentColor!Goldenrod!50}{#1}
        \else
            \pgfmathsetmacro{\PercentColor}{max(min(100.0*(\MidNumber - #1)/(\MidNumber-\MinNumber),100.0),0.00)} %
            \hspace{-0.33em}\colorbox{Red!\PercentColor!Goldenrod!50}{#1}
        \fi
}

\newcolumntype{R}{>{\collectcell\ApplyGradient}c<{\endcollectcell}}

\newcommand{\logo}[0]{\raisebox{-.25\height}{\includegraphics[width=.05\textwidth]{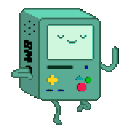}}} 

\makeatletter
\newcommand*{\radiobutton}{%
  \@ifstar{\@radiobutton0}{\@radiobutton1}%
}
\newcommand*{\@radiobutton}[1]{%
  \begin{tikzpicture}
    \pgfmathsetlengthmacro\radius{height("X")/3}
    \draw[radius=\radius] circle;
    \ifcase#1 \fill[radius=.6*\radius] circle;\fi
  \end{tikzpicture}%
}
\makeatother

\title{\logo \hspace{0.003em} Beemo: Benchmark of Expert-edited Machine-generated Outputs}

\input{sections/notations}

\begin{document}
\maketitle

\begin{abstract}
The rapid proliferation of large language models (LLMs) has increased the volume of machine-generated texts (MGTs) and blurred text authorship in various domains. However, most existing MGT benchmarks include \emph{single-author} texts (human-written \& machine-generated). This conventional design fails to capture more practical \emph{multi-author} scenarios, where the user refines the LLM response for natural flow, coherence, and factual correctness. Our paper introduces the \textbf{B}enchmark of \textbf{E}xpert-\textbf{e}dited \textbf{M}achine-generated \textbf{O}utputs (Beemo), which includes 6.5k texts written by humans, generated by ten instruction-finetuned LLMs and edited by experts for various use cases, ranging from creative writing to summarization. Beemo additionally comprises 13.1k machine-generated \& LLM-edited texts, which allows for diverse MGT detection evaluation across various edit types. We document the Beemo's creation protocol and present the results of benchmarking 33 configurations of MGT detectors in different experimental setups. We find that expert-based editing evades MGT detection, while LLM-edited texts are unlikely to be recognized as human-written. Beemo and all materials are publicly available.
\end{abstract}

\input{sections/introduction}

\input{tables/benchmarks}

\input{sections/related_work}

\input{sections/beemo}

\input{tables/exp_results_c}

\section{Experimental Setup}
\label{sec:experimental_setup}

\input{sections/detectors}

\input{sections/task_formulations}

\input{sections/performance_metrics}

\section{Results \& Analysis}
\input{sections/results}

\section{Conclusion \& Future Work}

\input{sections/conclusion}

\input{sections/limitations}

\input{sections/ethics}

\input{sections/acknowledgement}

\bibliography{anthology,custom}

\newpage
\clearpage

\appendix
\input{sections/appendix_annotators}
\clearpage
\newpage

\input{sections/appendix_guidelines}

\clearpage
\newpage

\input{sections/appendix_issues}
\clearpage
\newpage

\input{sections/appendix_LLM_edits}

\clearpage
\newpage

\input{sections/appendix_analysis}

\clearpage
\newpage

\input{sections/appendix_setup}

\clearpage
\newpage

\input{sections/appendix_results}

\end{document}

%% file: sections/notations.tex
\def\zephyr{\texttt{zephyr-7b-beta}\xspace}

\def\tulus{\texttt{tulu2-7b}\xspace}
\def\tulum{\texttt{tulu2-13b}\xspace}

\def\gemmas{\texttt{gemma-2b-it}\xspace}
\def\gemmam{\texttt{gemma-7b-it}\xspace}

\def\llamas{\texttt{Llama2-7b-chat-hf}\xspace}
\def\llamam{\texttt{Llama2-13b-chat-hf}\xspace}
\def\llamal{\texttt{Llama2-70b-chat-hf}\xspace}
\def\llamaxl{\texttt{Llama3.1-70B-Instruct}\xspace}

\def\gpt{\texttt{GPT-4o}\xspace}

\def\mistral{\texttt{Mistral-7B-Instruct-v0.1}\xspace}
\def\mixtral{\texttt{Mixtral-8x7B-Instruct-v0.1}\xspace}

%% file: sections/introduction.tex
\section{Introduction} 
The rapid advancement of large language models (LLMs) has significantly improved their ability to assist users with a wide range of writing tasks \cite{gero2022sparks,yang2022ai,chakrabarty2024creativity}. While the benefits of LLMs are commendable, their widespread adoption has raised concerns regarding authenticity of textual content and potential malicious uses within the domains of news, social media, science, and education \cite{lucas2023fighting,crothers2023machine,gupta2024sociodemographic,chamezopoulos-etal-2024-overview,tang2024science}. 

Over the last few years, a broad range of machine-generated text (MGT) detection benchmarks have been created to facilitate the development of reliable detectors, aimed at mitigating the risks associated with the misuse of LLMs across different domains and languages \citep[e.g.,][]{macko-etal-2023-multitude,macko2024multisocial,pu2023deepfake,dugan2024raid,nafis2024theseus,wang-etal-2024-m4,wang-etal-2024-m4gt}. However, most of them focus on a \emph{single-author} scenario, comprising only machine-generated and human-written texts. This well-established design does not account for common applications of LLMs, where the user refines the LLM's response for natural flow, coherence, and factual correctness. 

This paper introduces the \textbf{B}enchmark of \textbf{E}xpert-\textbf{e}dited \textbf{M}achine-generated \textbf{O}utputs (Beemo\footnote{Our benchmark is named after BMO (abbreviated from ``Be MOre'', phonetically spelled ``Beemo''), one of the main characters of Adventure Time. Logo: \href{https://slackmojis.com/emojis/67173-bmo}{\texttt{slackmojis.com/bmo}}.}), which consists of 6.5k texts written by humans, generated by ten open-source instruction-finetuned LLMs and edited by expert annotators, who are well-experienced in refining LLM-generated content. Furthermore, each MGT is edited by two state-of-the-art LLMs using several diverse editing prompts, which results in 13.1k machine-generated \& LLM-edited texts. Beemo covers five use cases: open-ended generation, rewriting, summarization, and open \& closed question answering (QA). Our design enables various diagnostic evaluation scenarios, ranging from out-of-domain MGT detection to analyzing how an MTG detector's behavior changes after the LLM response is refined by experts and ``humanized'' by other LLMs.

Our main contributions are: \begin{inparaenum}[(i)] \item we create Beemo, one of the first \emph{multi-author} benchmarks of LLM-generated \& expert-edited responses for fine-grained MGT detection, which counts 19.6k texts in total; \item we evaluate 33 configurations of zero-shot and pretrained MGT detectors; \item we release Beemo\footnote{\href{https://huggingface.co/datasets/toloka/beemo}{\texttt{hf.co/datasets/toloka/beemo}}} and all annotation materials\footnote{\href{https://github.com/Toloka/beemo}{\texttt{github.com/Toloka/beemo}}}. \end{inparaenum}

%% file: tables/benchmarks.tex
\begin{table*}[ht!]
    \centering
    \resizebox{\textwidth}{!}{
    \begin{tabular}{lrrcc}
    \toprule
      \textbf{Resource}   &  \textbf{Size} & \textbf{\# Models} & 
      \makecell{\textbf{Machine-generated} \\ \textbf{\& LLM-edited}} & \makecell{\textbf{Machine-generated} \\ \textbf{\& Human-edited}} \\
      \midrule
       TuringBench \cite{uchendu2021turingbench} &  200k & 19 & \xmark & \xmark \\
       RoFT \cite{dugan2023real} & 21k & 2 & \xmark & \xmark \\
       MULTITuDE \cite{macko-etal-2023-multitude} & 75k & 8 & \xmark & \xmark \\
       OpenLLMText \cite{chen2023token} & 340k & 5 & \xmark & \xmark \\
       HC3 Plus \cite{su2023hc3} &  210k & 1 & \xmark & \xmark \\
       MAGE \cite{li-etal-2024-mage} &  447k  & 27  & \xmark & \xmark\\
       RAID \cite{dugan2024raid} & 6.2M &  11 & \xmark & \xmark\\
       MultiSocial \cite{macko2024multisocial} & 472k & 7 & \xmark & \xmark \\
       BUST \cite{cornelius2024bust} & 25k & 7 & \xmark & \xmark \\
       M4 \cite{wang-etal-2024-m4} & 122k & 6 & \xmark & \xmark\\
       M4GT-Bench \cite{wang-etal-2024-m4gt} & 217k & 8 & \xmark & \xmark\\
       LLM-DetectAIve \cite{abassy2024llm} & 303k & 14 & \cmark & \xmark \\
       MixSet \cite{zhang-etal-2024-llm} & 3.6k & 8 & \cmark & \cmark \\
       LAMP \cite{chakrabarty2024can} & 1k & 7 & \cmark & \cmark \\
       \cdashline{1-5} \vspace{-1em}\\
       \textbf{Beemo (ours)} & 19.6k & 10 & \cmark & \cmark \\
       \bottomrule
    \end{tabular}}
    \caption{Comparison of publicly available monolingual and cross-lingual MGT detection evaluation resources. Beemo is one of the first benchmarks that contains machine-generated \& LLM/human-edited texts.}
    \label{tab:benchmarks}
\end{table*}

%% file: sections/related_work.tex
\section{Related Work}
\autoref{tab:benchmarks} summarizes commonly used monolingual and cross-lingual MGT detection evaluation resources. To the best of our knowledge, only three of them comprise machine-generated \& LLM-edited texts, and only two of them include machine-generated \& human-edited texts. Below, we provide an overview of the resources by their type and task formulation with the main focus on English.

\subsection{Standard MGT Detection Benchmarks}
MGT detection features various task formulations and labelling schemes: binary classification, neural authorship attribution, boundary-detection, and multi-author classification.

\paragraph{Binary Classification} Binary classification is a well-established design of MGT benchmarks \cite{radford2018improving,fagni2020tweepfake,liyanage2022benchmark,macko-etal-2023-multitude,cui2023said}. The task is to determine if a given text is machine-generated or not. 

\paragraph{Authorship Attribution} Authorship Attribution aims to identify the author of a given text \cite{uchendu2020authorship,uchendu2021turingbench,uchendu2023attribution}, a multi-class classification task with humans and LLMs as the  labels. 

\paragraph{Boundary Detection} Boundary detection is a less explored task formulation, which aligns with application of LLMs for text continuation tasks. Here, the goal is to detect a change point in the text, where a natural text transitions into a neural one \cite{dugan2023real,wang-etal-2024-m4gt}.

\begin{figure*}[ht!]
    \centering
    \includegraphics[width=\textwidth]{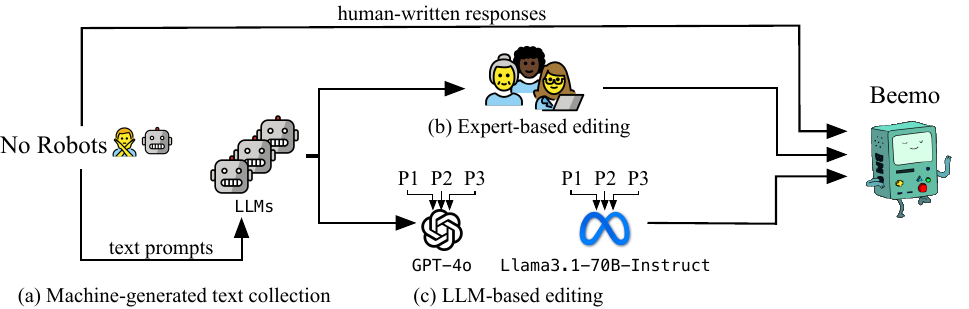}
    \caption{Overview of the Beemo's creation pipeline. (a) Use \texttt{No Robots} \cite{no_robots} as the source of prompts and human-written responses across five categories. Generate responses from ten open-source instruction-finetuned LLMs. (b) Refine the LLMs' responses with a team of expert editors. (c) Refine the LLMs' responses using two state-of-the-art LLMs and editing prompts (P1-P3). Each of 2,187 instances includes nine text versions.}
    \label{fig:beemo_flow}
\end{figure*}

\paragraph{Multi-author Classification} Recent research has proposed several fine-grained MGT benchmarks, which help to explore how varying degrees of LLM intervention in writing tasks affect the behavior of MGT detectors. MixSet \cite{zhang-etal-2024-llm} comprises human-written, machine-generated, and human/LLM-refined MGTs and focuses on multi-author binary classification. LLM-DetectAIve \cite{abassy2024llm} offers a four-way classification task with two other classes (human-written/machine-generated \& machine-polished), which reflects a common usage of LLMs for enhancing a human-written text. LAMP \cite{chakrabarty2024can} consists of 1k non- and fiction MGTs edited by professionals and LLMs and explores automatic detection and rewriting of ``problematic'' spans in MGTs. Our work differs from these studies in the following aspects: (i) in line with \citeauthor{zhang-etal-2024-llm,chakrabarty2024can}, we present one of the first attempts to create a benchmark of expert-edited MGTs; (ii) we use an instruction-tuning dataset as the source of prompts and human-written responses, which ensures coverage of various domains and use cases; (iii) our annotation protocol is not based on a predefined taxonomy of operations (MixSet) and edits (LAMP); instead, our annotators make the edits based on their expertise in refining LLM-generated content; (iv) Beemo includes the largest number of expert-edited MGTs compared to MixSet (1.2k) and LAMP (1k); (v) we analyze how the behavior of binary MGT detectors changes after edits are made  w.r.t. edit ratio and use case.

\subsection{Editing MGTs}
There are several approaches to editing generated content within MGT detection: operation-based, prompt-based, and expert-based editing. 

\paragraph{Operation-based Editing} \citet{zhang-etal-2024-llm} propose five operations to refine an MGT at the token-, sentence-, and paragraph-level: polish, complete, rewrite, humanize, and adapt. This hybrid editing approach emulates real-world cases where humans aim to modify MGT to suit their preferences, improve quality, or align with the intended purpose.

\paragraph{Prompt-based Editing} Emerging research leverages prompt-based LLM editing to mitigate the risks of ``humanizing'' LLM-generated content at scale. \citet{hu2023radar} employ few-shot in-context learning with human demonstrations to enhance MGT edits. \citet{mitchell2023detectgpt, yangdna} use prompts to simulate humans limited editing behavior using T5 \cite{raffel2020exploring}.  \citet{abassy2024llm} utilize various prompts to refine MGTs and polish human-written texts, such as improving grammar and fluency. While prompt-based editing show benefits to assist humans, the output quality depends heavily on prompts and LLM editors, emphasizing the importance of careful prompt design and LLM considerations \citep{kamalloo-etal-2023-evaluating,zhang2023prompting,zhang-etal-2024-benchmarking}.

\paragraph{Expert-based Editing}
Prior work employed human experts to edit and evaluate text across different domains \cite{nahar2024fakes, reid2022learning, du2022understanding, roberts1983evaluation, lucas2023fighting}. However, limited research utilizes such experts for editing MGTs. The above mentioned works by \citet{zhang-etal-2024-llm,chakrabarty2024can} highlight human experts' unique value in refining machine-generated content, particularly in adapting to academic genres. Our work incorporates the prompt- and expert-based editing approaches, summarizes the experts' strategies to refine LLM content, and presents the results of analyzing the detectors' performance w.r.t. edit percentages.

%% file: sections/beemo.tex
\input{tables/llms}

\section{Beemo}
\label{sec:beemo}
\autoref{fig:beemo_flow} outlines our high-level methodology for creating Beemo, which includes the following stages: generating instruction-finetuned LLMs' responses (\S\ref{subsec:generation}), editing the responses by expert annotators (\S\ref{subsec:expert_editing}) and state-of-the-art LLMs (\S\ref{subsec:llm_editing}). 

\subsection{Machine-generated Data Collection}
\label{subsec:generation}
\texttt{No Robots} \cite{no_robots} is a human-created instruction-finetuning dataset, which is used as the source of prompts and corresponding human-written responses across the following categories: open-ended generation, rewriting, summarization, and open and closed QA\footnote{We aim to select more general and practical categories, where the user is likely to refine the model response. We leave extending Beemo with other user-oriented categories for future work (e.g., chat and brainstorming).}. We randomly sample each prompt to generate a response with one of ten open-source instruction-finetuned LMs (see \autoref{tab:lms}), which range in size from 2B to 70B. We use the default HuggingFace \cite{wolf-etal-2020-transformers} chat templates and inference hyperparameters.

\subsection{Expert-based Editing}
\label{subsec:expert_editing}
We run an in-house annotation to create expert-edited versions of the MGTs. Our team consists of two lead editors and 25 expert annotators, who are native English speakers and well-experienced in refining content produced by LLMs\footnote{Our annotators are not experts in NLP but are well-experienced in annotating content generated by LLMs. Our annotators have diverse backgrounds, including professional writing, editing, and translation across various domains, from media and communication to education and technologies.}. Refer to a voluntary survey results from 17 respondents in \autoref{tab:labeler_demographics} (see \autoref{app:annotator_profiles}) for sociodemographic details. The lead editors collaborate closely with the annotators throughout the annotation project, suggesting areas for improvement, exchanging feedback in a group chat, and manually validating each annotator's submission. We provide detailed annotation guidelines to the annotators before they start the editing phase (see \autoref{section:appendix_guidelines}). Each annotator receives one example at a time (a category, a prompt, and an LLM's response) and is asked to (i) carefully read the prompt and the response; (ii) judge the response's relevance to the prompt and the category; (iii) fact check the response if required; (iv) edit the response by correcting factual inconsistency, removing hallucinations, and improving style, coherence, and fluency; and (v) proofread the edited version before submission. The recommended ratio of edits ranges between 20\% and 40\% of the response. The average pay rate is \$20/hr. The annotator can skip an example (i) if it does not require any edits and aligns with the prompt intent or (ii) if it requires significant improvements or does not follow the prompt closely. We discard such examples and use only the refined and manually validated responses to create Beemo.

\subsection{LLM-based Editing}
\label{subsec:llm_editing}
We utilize two open-source and proprietary LLMs for LLM-based editing: \llamaxl and \gpt (see \autoref{tab:lms}). We automatically refine the MGTs using three types of prompts based on distinct motivations (see \autoref{tab:chick-fil-a-values-comparison} in \autoref{appendix:llm_editing}): \begin{inparaenum}[(P1)] \item Focuses on grammatical correctness and human-like qualities; \item Aims to remove artifacts in an LLM-generated text (e.g., unwanted text such as \textit{''Sure, here is the summarized text''}); \item Emphasizes producing a more natural and native-sounding text. \end{inparaenum} Our aim here is to address the diversity of LLM-based edits in real-world scenarios, which target various aspects of making a machine-generated text more human-like \cite{zhang-etal-2024-llm,gero2022sparks}. We specify the recommended range of edits in (P1) and (P2) as 20\%--40\% and do not provide it in (P3) to ensure the LLM editors are not restricted in refining the generated response. 

\input{tables/stats_cat}

\subsection{Benchmark Analysis}
\label{subsec:bench_analysis}
\paragraph{General Statistics} 
We summarize the Beemo's general statistics by category (see \autoref{tab:stats_cat}) and LLM (see \autoref{tab:stats_model}, \autoref{tab:bertscore} in Appendix \ref{appendix:analysis}) based on count, stylometric, edit distance, and embedding-based similarity metrics computed via \texttt{TextDescriptives} \cite{hansen2023textdescriptives}, \texttt{editdistance}\footnote{\href{https://github.com/roy-ht/editdistance}{\texttt{github.com/roy-ht/editdistance}}} and \texttt{Evaluate}\footnote{\href{https://github.com/huggingface/evaluate}{\texttt{github.com/huggingface/evaluate}}}: the average text length in tokens, the average number of stopwords in a text, the Flesch-Kincaid grade of a text (FKG), Levenshtein distance (LD), and BERTScore similarity \cite{zhangbertscore} between different text versions.

We observe that the number of stopwords as a distinctive measure of repetitiveness \cite{frohling2021feature} depends on the LLM rather than its size. However, larger LLMs generally produce text with higher readability scores (FKG), with the exception of the \texttt{Llama2} LLMs. Analyzing the statistics by category, we find that the LLMs' responses and their expert-edited versions are generally longer than the human-written and LLM-edited texts. This implies that the LLM editors tend to shorten the generated responses, which is indicated by higher LD values. The average LD between the machine- and human-written responses ranges from 153 to 256, which suggests a minimal overlap between \texttt{No Robots} and the LLMs' instruction-finetuning corpora. The similarity between machine-generated texts and their edited versions remains consistent across LLMs based on BERTScore values. \gpt edits are the closest to the original machine-generated texts, while \llamaxl edits show lower similarity, reflecting diverse editing styles among LLM editors.

\begin{figure}[t!]
    \centering
    \includegraphics[width=\columnwidth]{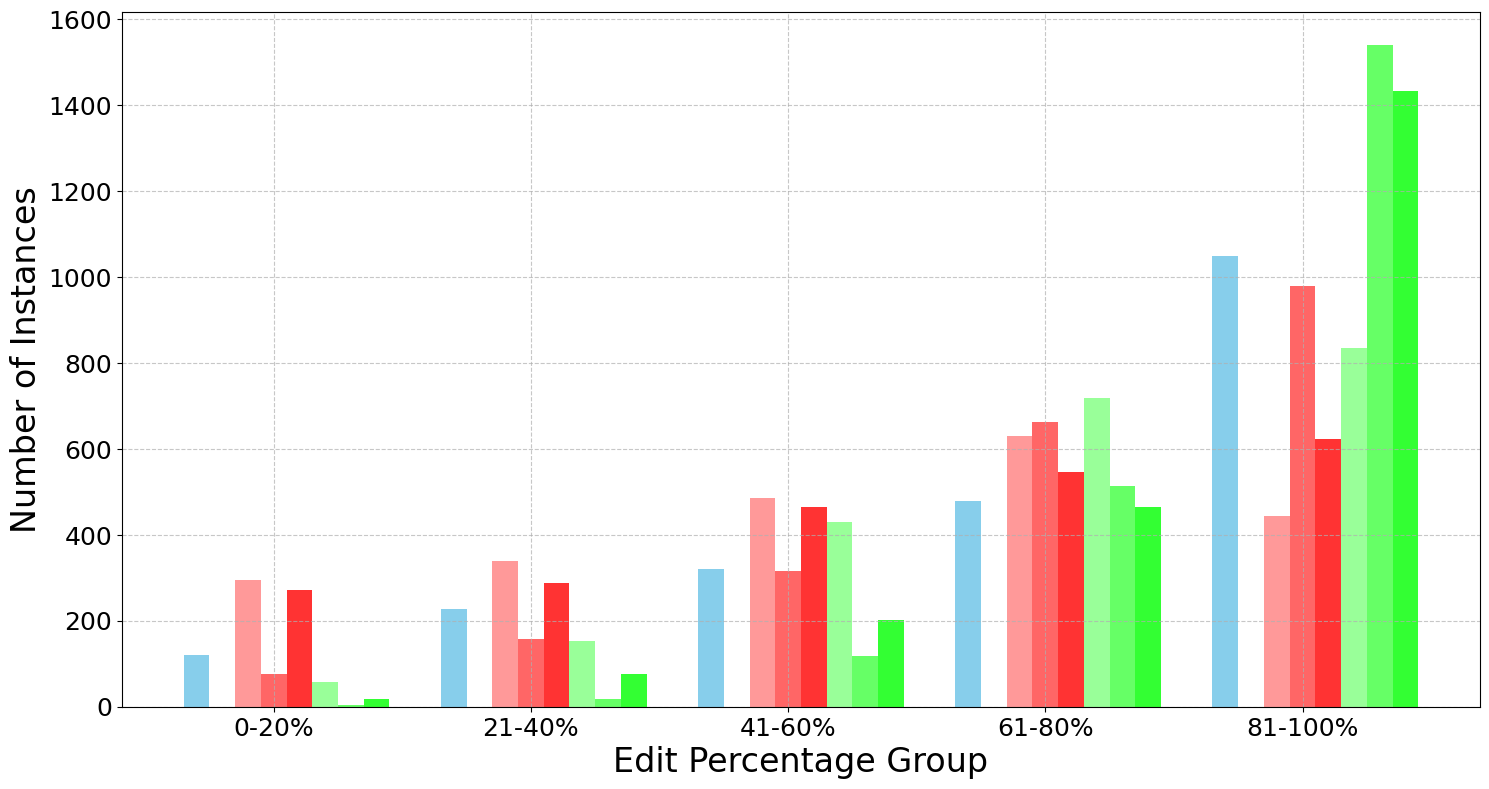}
    \caption{Distribution of edit percentages across five edit ranges for \textcolor{blue!60}{expert annotators}, \textcolor{red!100}{\gpt} (\textcolor{red!40}{\textbf{P1}}, \textcolor{red!70}{\textbf{P2}}, \textcolor{red!100}{\textbf{P3}}), and \textcolor{green!100}{\llamaxl} (\textcolor{green!40}{\textbf{P1}}, \textcolor{green!70}{\textbf{P2}}, \textcolor{green!100}{\textbf{P3}}). The bars represent the number of instances falling within each edit percentage range for each editor type.}
    \label{fig:edit-distance-distribution}
\end{figure}

\begin{figure}[t!]
    \centering
    \includegraphics[width=\columnwidth]{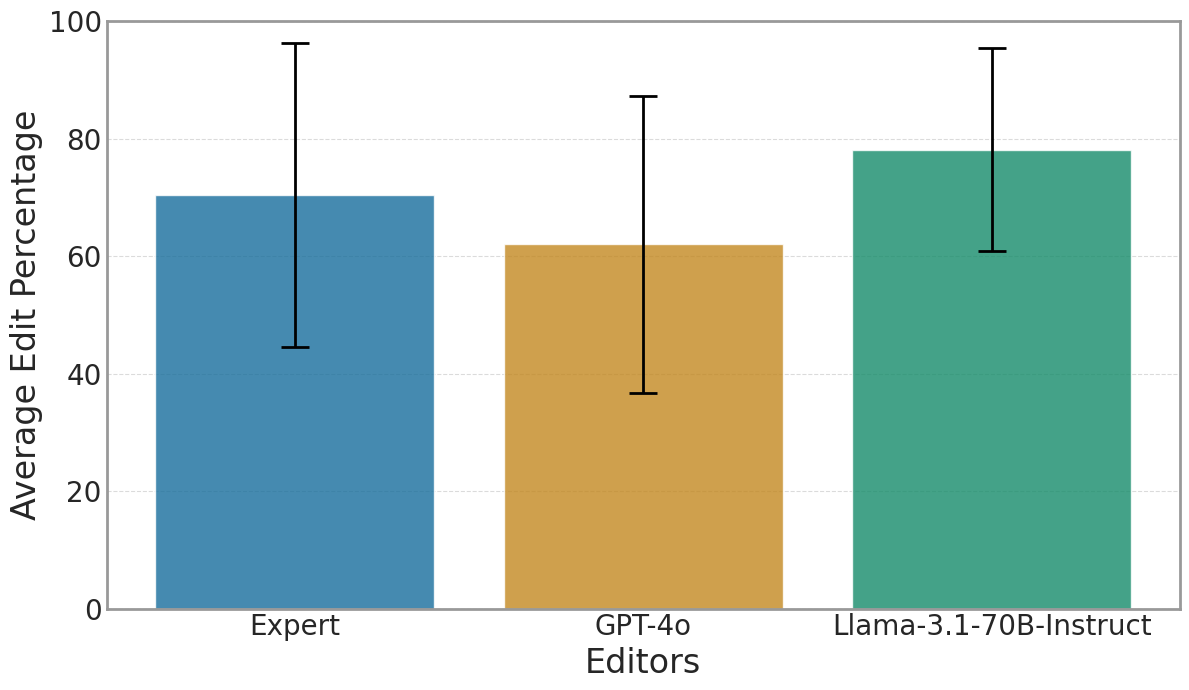}
    \caption{Comparison of average edit percentages among expert editors, \gpt, and \llamaxl. The bars represent the mean edit percentage for each editor type, with error bars indicating the standard deviation.}
    \label{fig:human-llm-edits}
\end{figure}

\paragraph{Editing Analysis} We compare the edit percentages between the MGTs and their  \llamaxl-, \gpt- and expert-edited versions using \texttt{difflib}\footnote{\href{https://docs.python.org/3/library/difflib.html}{\texttt{docs.python.org/library/difflib}}}. \autoref{fig:edit-distance-distribution} and
\autoref{fig:human-llm-edits}    present a comparison of editing behaviors among the expert and LLM editors. The experts demonstrate an average edit percentage of 70\%, falling between \gpt (60\%) and \llamaxl (80\%). The error bars indicate considerable variability in edit percentages across different instances. This variability is particularly pronounced for the LLM editors, indicating less consistency in their editing behavior than experts. Both LLMs make more extensive edits than specified in the prompts (\S\ref{subsec:llm_editing}), which suggests their potential limitation in controlled editing.

\paragraph{Effect of Prompt} \autoref{fig:prompt-comparison} (see \autoref{appendix:analysis}) illustrates the LLMs' editing behavior across the three prompts. Overall, \llamaxl makes the most extensive edits irrespective of the prompt. Both LLMs exhibit an increase in edit percentage from (P1) to (P2) of up to 15\%, although (P2) aims to control the edit range. Using (P3) results in a slight decrease in edit percentages for both LLMs compared to (P2), but they remain higher than (P1). The results suggest that prompt engineering significantly influences the LLM's ability to follow specified editing ranges, which aligns with \citet{wang2023element}.

\paragraph{Survey on Editing Strategies} We send a voluntary survey to our editors to better understand their main editing strategies and identify common issues in the generated responses (see \autoref{appendix:survey}). Our analysis of 14  respondents' answers reveals several common issues with the MGTs: (i) format inconsistencies (e.g., a partial or inconsistent ordering of items in a list); (ii) hallucinations and factual errors; (iii) struggling with creative writing (e.g., writing poems and haikus); (iv) complicated and repetitive vocabulary, and their responses can feel over-explained; and (v) responses often lack natural flow and exhibit a repetitive structure. Our expert editors follow a consistent approach based on their individual experience to refine the generated outputs: (i) carefully reading the prompt and LLM's response; (ii) fact-checking if required; (iii) focusing on complicated use of passive voice, repeated elements, and odd openings as common properties in our data; (iv) editing for natural flow, improving structure, and changing vocabulary for clarity and richness; (v) ensuring the response adheres to the required format; (vi) double-checking for grammar, phrasing and spelling errors; and (vii) reading through or aloud for a final quality check. Refer to \autoref{tab:common_issues_examples} in \autoref{appendix:issues} for examples of identified issues in machine-generated texts and corresponding edits.  Our experts report that it can be challenging to follow the recommended edit range of 20\%--40\%, and rewriting entire sections can be more effective than editing specific parts, particularly for summarization, rewriting, and open-ended generation. However, the core content of the LLM response remains consistent, even at higher edit ratios. 

%% file: tables/llms.tex
\begin{table*}[htp!]
    \centering
    \scriptsize
    \resizebox{\textwidth}{!}{ %
    \begin{tabular}{lllrc}
    \toprule
    \textbf{Model} & \textbf{Base} & \textbf{License} & \textbf{Source} & \textbf{SFT Corpus} \\
    \midrule
    \multicolumn{5}{c}{\textbf{Data generation}} \\
    \midrule

    \href{ https://huggingface.co/HuggingFaceH4/zephyr-7b-beta}{\zephyr} & Mistral-7B-v0.1 
    & MIT & \citet{tunstall2023zephyr} 
    & \makecell{UltraChat, \\ UltraFeedback } \\ 
    
    \vspace{-1em}\\ \cdashline{1-5} \vspace{-1em}\\

    \href{https://huggingface.co/allenai/tulu-2-7b}{\tulus} 
    & Llama 2 7B & \multirow{2}{*}{AI2 ImpACT} & \multirow{2}{*}{\citet{ivison2023camels}} & \multirow{2}{*}{\makecell{human-created, \\ synthetic}} \\ 
    
    \href{https://huggingface.co/allenai/tulu-2-13b}{\tulum} 
    & Llama 2 13B  &  & &  \\

    \vspace{-1em}\\ \cdashline{1-5} \vspace{-1em}\\

    \href{https://huggingface.co/google/gemma-2b-it}{\gemmas} 
    &  Gemma 2B & \multirow{2}{*}{Gemma} & \multirow{2}{*}{\citet{team2024gemma}} & \multirow{2}{*}{\makecell{human-created, \\ synthetic}} \\
    \href{https://huggingface.co/google/gemma-7b-it}{\gemmam} 
    & Gemma 7B & & & \\

    \vspace{-1em}\\ \cdashline{1-5} \vspace{-1em}\\

    \href{https://huggingface.co/meta-llama/Llama-2-7b}{\llamas} 
    & Llama 2 7B & \multirow{3}{*}{Llama} & \multirow{3}{*}{\citet{touvron2023llama}} & \multirow{3}{*}{Misc.} \\
    \href{https://huggingface.co/meta-llama/Llama-2-13b}{\llamam} 
    & Llama 2 13B & &  \\
    \href{https://huggingface.co/meta-llama/Llama-2-70b}{\llamal} 
    & Llama 2 70B & & \\

    \vspace{-1em}\\ \cdashline{1-5} \vspace{-1em}\\
    \href{https://huggingface.co/mistralai/Mistral-7B-v0.1}{\mistral} & Mistral-7B-v0.1 & Apache-2.0 & 
    \citet{jiang2023mistral} & Misc. \\

    \vspace{-1em}\\ \cdashline{1-5} \vspace{-1em}\\
    \href{https://huggingface.co/mistralai/Mixtral-8x7B-Instruct-v0.1}{\mixtral} 
    & Mixtral 8x7B & Apache-2.0 & \citet{jiang2024mixtral} &  Misc. \\

    \midrule
    \multicolumn{5}{c}{\textbf{LLM-based editing}} \\
    \midrule
    \href{https://huggingface.co/meta-llama/Llama-3.1-70B-Instruct}{\llamaxl} 
    & Llama 3.1 70B & Llama & \citet{dubey2024llama} & Misc. \\

    \vspace{-1em}\\ \cdashline{1-5} \vspace{-1em}\\
    
    \href{https://openai.com/index/gpt-4o-mini-advancing-cost-efficient-intelligence/}{\gpt} 
    & GPT-4 & OpenAI & \citet{openai2024gpt4omini} & Misc. \\

    \bottomrule
    \end{tabular}
    } 
    \caption{The LLMs used to generate (\S\ref{subsec:generation}) and edit (\S\ref{subsec:llm_editing}) responses. The supervised finetuning (SFT) / instruction-tuning corpus description is based on public information. We evaluate potential overlap between the SFT corpora and LLMs' outputs in \S\ref{subsec:bench_analysis}. Corpora references: UltraChat \cite{ding2023enhancing}; UltraFeedback \cite{cui2023ultrafeedback}.}
    \label{tab:lms}
\end{table*}

%% file: tables/stats_cat.tex
\begin{table*}[ht!]
    \centering
    \resizebox{\textwidth}{!}{
\begin{tabular}{lcrrrrrr}
\toprule
\textbf{Category} & \textbf{\# Examples} & \textbf{\# Tokens (P)} & \textbf{\# Tokens (H)} & \textbf{\# Tokens (M)} & \textbf{\# Tokens (E)} & \textbf{\# Tokens (L)} & \textbf{\# Tokens (G)} \\
\midrule
\texttt{Closed QA} & 1,845 & 268.0 \small$\pm$ 230.3 & 26.7 \small$\pm$ 17.4 & 83.3 \small$\pm$ 73.2 & 60.9 \small$\pm$ 66.4 & 74.2 \small$\pm$ 35.5 & 62.4 \small$\pm$ 37.4 \\
\texttt{Generation} & 5,265 & 40.3 \small$\pm$ 34.3 & 225.5 \small$\pm$ 142.9 & 305.6 \small$\pm$ 175.4 & 278.1 \small$\pm$ 171.4 & 115.8 \small$\pm$ 23.1 & 115.9 \small$\pm$ 28.5 \\
\texttt{Open QA} & 4,347 & 15.4 \small$\pm$ 44.9 & 89.1 \small$\pm$ 55.0 & 186.3 \small$\pm$ 145.4 & 128.4 \small$\pm$ 121.5 & 100.7 \small$\pm$ 30.2 & 91.7 \small$\pm$ 37.7 \\
\texttt{Rewrite} & 4,725 & 296.7 \small$\pm$ 249.7 & 240.6 \small$\pm$ 201.0 & 250.3 \small$\pm$ 160.2 & 242.7 \small$\pm$ 170.3 & 115.7 \small$\pm$ 22.1 & 112.2 \small$\pm$ 28.2 \\
\texttt{Summarize} & 3,501 & 274.9 \small$\pm$ 162.9 & 73.7 \small$\pm$ 45.2 & 143.4 \small$\pm$ 110.6 & 98.2 \small$\pm$ 59.9 & 102.5 \small$\pm$ 25.0 & 85.9 \small$\pm$ 30.8 \\
\vspace{-1em}\\ \cdashline{1-8} \vspace{-1em}\\
\texttt{Overall} & 19,683 & 159.4 \small$\pm$ 204.0 & 153.3 \small$\pm$ 151.6 & 216.3 \small$\pm$ 164.0 & 184.2 \small$\pm$ 160.4 & 101.8 \small$\pm$ 27.2 & 93.6 \small$\pm$ 32.5 \\
\bottomrule
\end{tabular}
}
    \caption{General statistics by category. \textbf{P}=\texttt{No Robots} prompts (2,187 prompts); \textbf{H}=human-written (2,187 texts); \textbf{M}=machine-generated (2,187 texts); \textbf{E}=expert-edited (2,187 texts); \textbf{L}=\llamaxl-edited (6,561 texts); \textbf{G}=\gpt-edited (6,561 texts). \textbf{L} and \textbf{G} are aggregated over three editing prompts (\S\ref{subsec:llm_editing}).}
    \label{tab:stats_cat}
\end{table*}

%% file: tables/exp_results_c.tex
\begin{table*}[ht!]
\centering
\setlength{\tabcolsep}{10pt}
{\renewcommand{\arraystretch}{1}}
\resizebox{\textwidth}{!}{
\begin{tabular}{llRRRRRRR}
\toprule
\multicolumn{2}{p{2ex}}{\textbf{Detector}} & \multicolumn{1}{c}{\textbf{H vs. M}} & \multicolumn{1}{c}{\textbf{E vs. M}} & \multicolumn{1}{c}{\textbf{H vs. E}} & \multicolumn{1}{c}{\textbf{L vs. M}} & \multicolumn{1}{c}{\textbf{H vs. L}} & \multicolumn{1}{c}{\textbf{G vs. M}} & \multicolumn{1}{c}{\textbf{H vs. G}} \\
\midrule

\multicolumn{9}{c}{\textbf{Zero-shot MGT Detectors}} \\ \midrule

\multicolumn{2}{l}{\textbf{Binoculars}} & 83.90 & 76.79 & 61.24 & 59.90 &  79.90  & 57.93 &  78.15 \\ \midrule

\multirow{4}{*}{\textbf{Log Probability}} 
& \small{\texttt{GPT2-XL}} & 69.72 & 64.73 & 56.56 & 63.70 & 58.88 & 60.32 & 61.23 \\ 

& \small{\texttt{OPT-1.3B}} & 74.82 & 66.52 & 60.86 & 66.58 & 62.78 & 61.12 & 66.95 \\
& \small{\texttt{Falcon-7B}} & 86.07 & 65.48 & 78.41 & 65.36 & 80.78 & 59.65 & 82.44 \\
& \small{\texttt{Qwen2-7B}} & 87.77 & 68.68 & 78.02 & 56.23 & 84.52 & 62.45 & 81.27 \\
\midrule 
\multirow{4}{*}{\textbf{Rank}} 
& \small{\texttt{GPT2-XL}} & 60.36 & 56.70 & 54.50 & 59.85 & 50.65 & 56.79 & 53.64 \\
& \small{\texttt{OPT-1.3B}} & 65.95 & 57.44 & 59.94 & 63.56 & 53.48 & 57.10 & 59.86 \\
& \small{\texttt{Falcon-7B}} & 74.52 & 56.56 & 70.66 & 62.95 & 63.64 & 55.57 & 70.61 \\
& \small{\texttt{Qwen2-7B}}& 74.22 & 57.65 & 69.11 & 52.55 & 72.52 & 55.10 & 70.81 \\
\midrule

\multirow{4}{*}{\textbf{Log-Rank}} 
& \small{\texttt{GPT2-XL}} & 70.26 & 64.95 & 56.78 & 65.25 & 57.42 & 61.20 & 60.68 \\
& \small{\texttt{OPT-1.3B}}& 73.03 & 66.11 & 59.03 & 67.21 & 59.30 & 61.83 & 63.90 \\
& \small{\texttt{Falcon-7B}} & 84.81 & 65.07 & 77.02 & 66.12 & 78.26 & 60.26 & 80.32 \\
& \small{\texttt{Qwen2-7B}} & 86.91 & 68.13 & 77.04 & 56.04 & 83.62 & 62.09 & 80.33 \\
\midrule

\multirow{4}{*}{\textbf{Entropy}} 
& \small{\texttt{GPT2-XL}} & 38.93 & 42.42 & 46.06 & 36.74 & 51.72 & 41.95 & 46.64 \\
& \small{\texttt{OPT-1.3B}} & 37.63 & 42.74 & 44.18 & 35.41 & 51.33 & 41.07 & 45.72 \\
& \small{\texttt{Falcon-7B}}& 20.92 & 41.65 & 24.42 & 36.02 & 29.24 & 40.96 & 25.66 \\
& \small{\texttt{Qwen2-7B}} & 10.36 & 32.02 & 17.74 & 44.01 & 12.82 & 38.01 & 15.28 \\
\midrule

\multirow{4}{*}{\makecell[l]{\textbf{DetectLLM} \\ \textbf{Likelihood Log-Rank Ratio}}} 
& \small{\texttt{GPT2-XL}} & 68.73 & 63.72 & 56.18 & 67.45 & 51.86 & 62.27 & 57.07 \\
& \small{\texttt{OPT-1.3B}} & 64.37 & 63.09 & 51.90 & 67.21 & 47.17 & 62.49 & 52.10 \\
& \small{\texttt{Falcon-7B}} & 76.73 & 62.16 & 68.41 & 66.77 & 64.68 & 61.28 & 68.31 \\
& \small{\texttt{Qwen2-7B}} & 79.68 & 64.12 & 69.40 & 54.71 & 76.25 & 59.41 & 72.83 \\
\midrule

\multirow{4}{*}{\makecell[l]{\textbf{DetectLLM} \\ \textbf{Normalized Perturbed Log-Rank}}} 
& \small{\texttt{GPT2-XL}} & 64.06 & 63.07 & 62.00 & 54.08 & 63.34 & 51.07 & 63.93 \\
& \small{\texttt{OPT-1.3B}} & 65.29 & 63.86 & 62.86 & 56.98 & 64.04 & 51.36 & 65.12 \\
& \small{\texttt{Falcon-7B}} & 70.14 & 63.18 & 66.71 & 56.65 & 68.40 & 51.37 & 69.84 \\
& \small{\texttt{Qwen2-7B}} & 74.12 & 68.30 & 68.11 & 54.52 & 72.49 & 50.68 & 73.84 \\
\midrule

\multirow{4}{*}{\textbf{DetectGPT}} 
& \small{\texttt{GPT2-XL}} & 64.91 & 63.71 & 62.75 & 59.15 & 63.53 & 58.97 & 63.63 \\
& \small{\texttt{OPT-1.3B}} & 67.27 & 66.87 & 64.22 & 64.39 & 64.68 & 61.65 & 65.32 \\
& \small{\texttt{Falcon-7B}} & 70.69 & 65.35 & 67.47 & 64.15 & 67.73 & 62.67 & 68.18 \\
& \small{\texttt{Qwen2-7B}} & 72.76 & 69.37 & 68.21 & 61.36 & 70.14 & 63.05 & 69.90 \\
\midrule

\multicolumn{9}{c}{\textbf{Pretrained MGT Detectors}} \\ \midrule

\textbf{RADAR} &  &  51.41 & 50.93 & 50.80 & 34.51 & 62.65 & 37.96 & 60.16 \\
\textbf{MAGE} & & 73.72 & 60.88 & 64.09 & 57.95 & 67.79 & 60.59 & 65.37 \\
\textbf{AIGC MPU} & & 70.52 & 70.35 & 50.66 & 65.71 & 57.17 & 59.66 & 62.69 \\
\textbf{OpenAI RoBERTa-base} & & 66.95 & 66.83 & 49.13 & 66.95 & 50.00 & 66.95 & 50.00 \\
\textbf{OpenAI RoBERTa-large} &  & 62.93 & 63.96 & 48.76 & 64.71 & 48.57 & 59.67 & 53.74 \\
\bottomrule
\end{tabular}
}
\caption{The AUROC scores (\%) of the 33 MGT detectors' configurations on Beemo. A random classifier has an AUROC of 50\%. \textbf{H}=human-written; \textbf{M}=machine-generated; \textbf{E}=expert-edited; \textbf{L}=\llamaxl-edited; \textbf{G}=\gpt-edited. \textbf{L} and \textbf{G} are aggregated over three prompts (\S\ref{subsec:llm_editing}).}
\label{tab:results}
\end{table*}

%% file: sections/detectors.tex
We use Beemo as an out-of-domain benchmark to evaluate generalization abilities of 33 zero-shot and pretrained MGT detectors' configurations in seven binary classification task formulations, which rely on single- and multi-author text versions. We list the detectors below and detail them in \autoref{appendix:setup}.

\paragraph{Zero-shot MGT Detectors} The zero-shot MGT detectors utilize log probabilities, entropy, and curvature-based properties to score a given text:  \begin{inparaenum}[(i)] \item Binoculars \cite{hansspotting}; \item Log probability \cite{solaiman2019release}; \item Rank \cite{gehrmann-etal-2019-gltr}; \item Log-Rank \cite{ippolito-etal-2020-automatic}; \item Entropy \cite{gehrmann-etal-2019-gltr,mitchell2023detectgpt}; \item DetectLLM  \cite{su-etal-2023-detectllm} Likelihood Log-Rank ratio (LRR) and Normalized Perturbed Log-Rank (NPR); and \item DetectGPT \cite{mitchell2023detectgpt}\footnote{DetectLLM NPR and DetectGPT rely on an additional LLM (\href{https://huggingface.co/google-t5/t5-3b}{\texttt{T5-3B}}) to perturb the input text and score both original and perturbed texts. The perturbation number is set to 20.}. \end{inparaenum} We use the codebase by \citeauthor{hansspotting} and \citeauthor{su-etal-2023-detectllm} to run the detectors, which supports \texttt{GPT2-XL}\footnote{\href{https://huggingface.co/openai-community/gpt2-xl}{\texttt{hf.co/openai-community/gpt2-xl}}}, \texttt{OPT-1.3B}\footnote{\href{https://huggingface.co/facebook/opt-1.3b}{\texttt{hf.co/facebook/opt-1.3b}}}, \texttt{Falcon-7B}\footnote{\href{https://huggingface.co/tiiuae/falcon-7b}{\href{https://huggingface.co/tiiuae/falcon-7b}{\texttt{hf.co/tiiuae/falcon-7b}}}}, and \texttt{Qwen2-7B}\footnote{\href{https://huggingface.co/Qwen/Qwen2-7B}{\texttt{hf.co/Qwen/Qwen2-7B}}} as the default backbone  models.

\paragraph{Pretrained MGT Detectors} We consider the following pretrained detectors: \begin{inparaenum}[(i)] \item 
RADAR\footnote{\href{https://huggingface.co/TrustSafeAI/RADAR-Vicuna-7B}{\texttt{hf.co/TrustSafeAI/RADAR-Vicuna-7B}}} \cite{radar}; \item AIGC MPU\footnote{\href{https://huggingface.co/yuchuantian/AIGC_detector_env1}{\texttt{hf.co/yuchuantian/AIGC\_detector\_env1}}}; \item MAGE\footnote{\href{https://huggingface.co/yaful/MAGE}{\texttt{hf.co/yaful/MAGE}}} \cite{li-etal-2024-mage}; and \item OpenAI RoBERTa-base\footnote{\href{https://huggingface.co/openai-community/roberta-base-openai-detector}{\texttt{hf.co/openai/roberta-base-openai-detector}}}/large\footnote{\href{https://huggingface.co/openai-community/roberta-large-openai-detector}{\texttt{hf.co/openai/roberta-large-openai-detector}}}.
\end{inparaenum}

%% file: sections/task_formulations.tex
\paragraph{Task Formulations} We explore two research questions in our work: (1) \textit{Do the detectors identify MGTs as generated after they are refined by an expert or by an LLM?} and (2) \textit{Do the detectors identify MGTs as human-written after they are refined by an expert or by an LLM?} We consider the corresponding binary classification task formulations by using different text versions as the input:
\begin{enumerate}[noitemsep,topsep=0.1pt,leftmargin=1.2em]
    \item human-written (\textbf{H}; \texttt{label=0}) vs. machine-generated, expert-, \llamaxl-, and \gpt-edited (\textbf{M/E/L/G}; \texttt{label=1}). We treat a text as machine-generated even if the experts edit it. 
    \item  expert-, \llamaxl-, and \gpt-edited (\textbf{E/L/G}; \texttt{label=0}) vs. machine-generated (\textbf{M}; \texttt{label=1}). We treat a text as human-written even if the LLMs ``humanize'' it.
\end{enumerate}

%% file: sections/performance_metrics.tex
\paragraph{Performance Metric} Following \citet{verma-etal-2024-ghostbuster,mitchell2023detectgpt}, we evaluate the detector performance using Area Under the Receiver Operating Characteristic curve (AUROC). AUROC values represent the probability that a detector assigns higher scores to a randomly selected machine-generated text than a randomly selected human-written text. AUROC is commonly used in zero-shot scenarios -- where the choice of the scoring threshold is crucial -- as it considers the range of all possible thresholds \cite{krishna2024paraphrasing}.

%% file: sections/results.tex
This section describes our empirical evaluation results on Beemo. We report the overall results in \autoref{tab:results} and results by category in \autoref{appendix:results}. Overall, we observe that Binoculars is the best-performing zero-shot detector, while MAGE is the strongest pretrained detector in most scenarios. Entropy performs worse or on par with a random guessing classifier. We summarize our findings below w.r.t. task formulation, editing type, MGT detector type, category, and edit percentages. 

\paragraph{Expert-based Editing Evades Detection} Analyzing the results in the \textbf{H vs. M/E/L/G} scenarios, we find that all detectors often fail to recognize an MGT after it is refined by experts, with the AUROC scores decreasing by up to 22\% for zero-shot (e.g., Binoculars, Log Probability, and Log-Rank) and pretrained (e.g., AIGC MPU, OpenAI RoBERTa-base/large) detectors. In contrast, many zero-shot detectors identify LLM-edited texts as MGTs, with Binoculars, Rank, Log-Rank, and Log Probability demonstrating stronger generalization abilities. However, RADAR, AIGC MPU, and OpenAI RoBERTa-base/large exhibit a random guessing performance across various edit types, which suggests a strong distribution shift.

\paragraph{LLM-edited Texts are Less Likely to Be Recognized as Human-authored} A comparison of the results in the \textbf{E/L/G vs. M} scenarios confirms our previous finding that the expert-edited texts are more likely to be classified as human-written compared to the LLM-edited ones. Notably, \gpt-edited texts are generally easier to identify as MGTs compared to expert- and \llamaxl-edited.

\paragraph{Analysis by Detector Type} Here, we compare the results between the zero-shot and pretrained detectors in more detail. The zero-shot detectors (i) better distinguish between human-written and machine-generated texts (\textbf{H vs. M}); (ii) are more generalizable to expert- and LLM-edited texts, which is indicated by lower AUROC $\delta$-scores (e.g., up to 10\% for Log-Rank, DetectLLM LLR/NPR, DetectGPT and up to 20\% for AIGC MPU, MAGE, and OpenAI RoBERTa-base/large); and (iii) perform better with a larger backbone model in the \textbf{H vs. M/E/L/G} scenarios, however, this effect is less pronounced in the \textbf{E/L/G vs. M} scenarios. 

\begin{figure}[t!]
    \centering
    \includegraphics[width=0.94\linewidth]{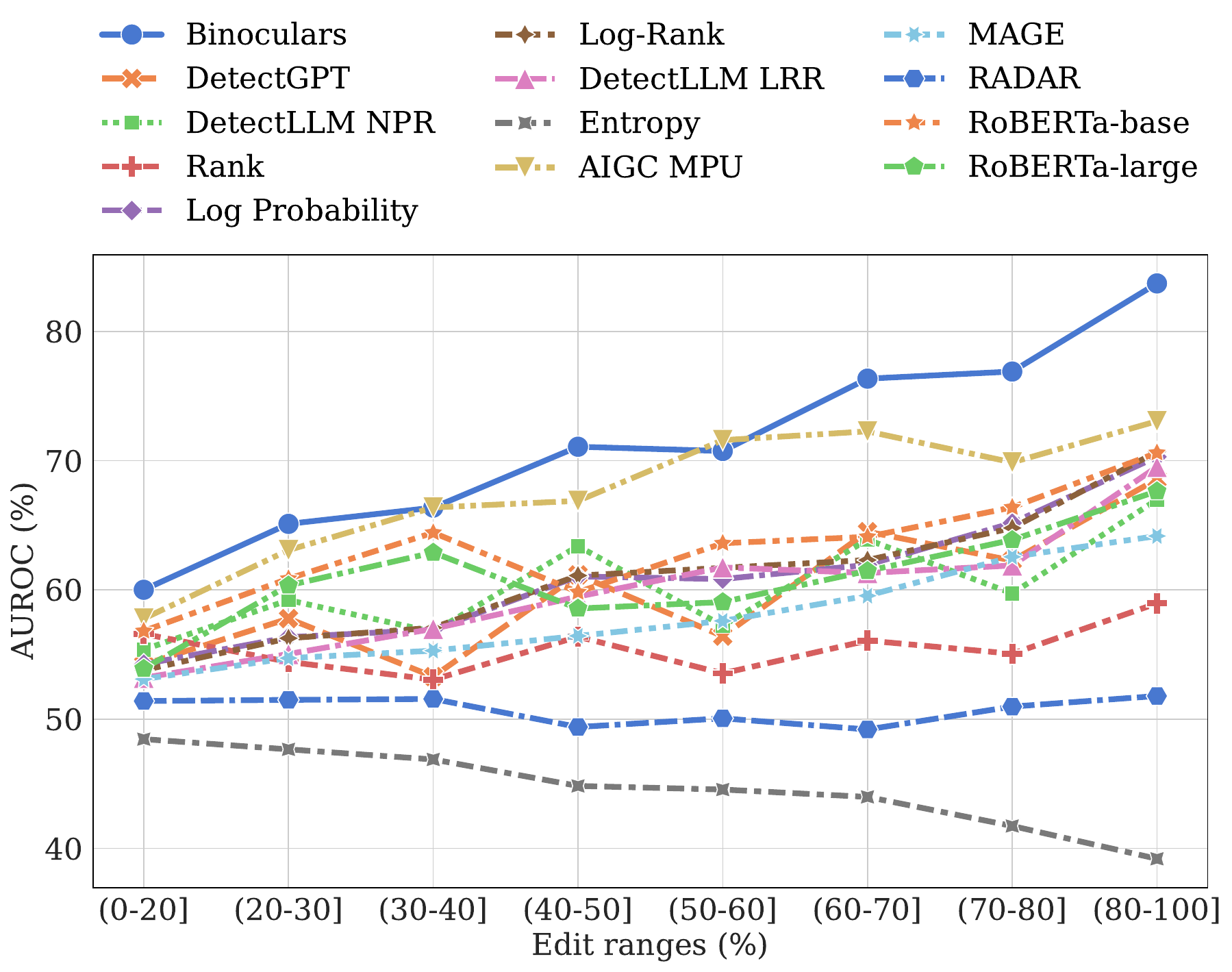}
    \caption{Results in the ``Expert-edited'' (\texttt{label=0}) vs. ``machine-generated'' (\texttt{label=1)} scenario divided into seven groups by the edit range.}
    \label{fig:human-ratio}
\end{figure}

\paragraph{Analysis by Category} The key finding here is that the detectors' behavior w.r.t. different edit types remains the same, but the AUROC scores across most task formulations depend on category (see \autoref{appendix:results}). The detectors perform consistently better on open-ended generation and open QA (e.g., Binoculars=up to 86\%; Log Probability=up to 93\%; DetectLLM LLR=up to 86\%; MAGE=up to 76\% in \textbf{H vs. M/E/L/G}). The rewriting, summarization, and closed QA are more challenging, where the detectors exhibit significant performance drops after expert edits (e.g., Log Probability=up to 20\%; Rank=up to 17\%; Log-Rank=up to 23\%; DetectLLM=up to 18\%) or struggle with distinguishing between different MGT versions (e.g., RADAR and MAGE often perform worse than a random classifier). We attribute these trends to category-specific text generation issues identified by our experts in \S\ref{sec:beemo} and shorter text length (see \autoref{tab:stats_cat}), which remains an unresolved challenge in MGT detection.

\paragraph{Effect of Edit Percentage} We analyze the effect of the edit percentage in the \textbf{E vs. M} scenario to assess the impact of expert edits (see \autoref{fig:human-ratio}). The overall trend is that the detectors' performance remains moderate and stable across the experts' edit ratios, showing general improvement as the ratio increases. Most detectors achieve similar AUROC scores for both moderately (20\%--40\%) and significantly (60\%--80\%) edited texts.  Lower AUROC scores at the lower edit ranges indicate that moderate editing can confuse the detectors.

%% file: sections/conclusion.tex
This work introduces Beemo, one of the first multi-author benchmarks of expert-edited and LLM-refined MGTs for English. Beemo covers five common use cases of instruction-finetuned LLMs, ranging from creative writing to summarization. We describe the Beemo creation approach and common issues in LLM responses and strategies for mitigating them based on the feedback from our expert editors. We conduct an extensive out-of-domain evaluation of 33 MGT detectors' configurations and analyze their performance in seven signle- and multi-author binary classification tasks. Our key empirical results demonstrate that detectors can be confused by moderate expert edits, while editing with state-of-the-art LLMs does not significantly influences the detection behavior. Furthermore, we find that zero-shot detectors are more generalizable to both expert- and LLM-edited MGTs than pretrained detectors. Our \emph{future} work includes (i) exploration of other MGT detection task formulations, including authorship attribution, k-way classification (human-written, machine-generated, machine-generated \& expert-/LLM-edited), and extraction of spans; (ii) establishment of the human baseline across various editing types; (iii) ablation studies on exploring the supervised detectors' robustness towards unseen instruction-finetuned LLM and writing task; and (iv) ablation studies on the effect of the LLM-edited MGTs in the training data on the detectors' generalization.

%% file: sections/limitations.tex
\section*{Limitations}
\paragraph{Evaluation Design} While we present the results of extensive empirical evaluation of a broad range of MGT detectors' configurations, we acknowledge certain limitations in our evaluation design. First, the LLM-based editing inherently depends on the prompt, and the best prompt configuration is LLM-specific \cite{voronov2024mind}. Although we address this sensitivity by employing three distinct types of editing prompts, our results may still be affected by the effect of prompt engineering. Second, our MGT detectors represent commonly used open-source approaches, which does not address the analysis of commercial and API-based MGT detectors often used to check if the LLM's response refinement passes them. Furthermore, ensemble approaches can be considered, which are more generalizable to the out-of-domain data \cite{wang-etal-2024-semeval-2024}. Last but not least, there might be a distribution shift with respect to how a non-expert user can refine their LLM responses in terms of editing quality. However, a direct comparison of expert and non-expert editors falls outside the scope of this work.

\paragraph{Need for Continuous Data Collection} The design of MGT benchmarks is fragile due to the rapid development of novel LLMs and their instruction-finetuned versions. This raises the need to continuously update Beemo to keep it with the current state of the field; however, it is expensive to collect expert-edited texts at scale. We encourage the NLP researchers and practitioners to contribute machine-generated and machine-generated \& LLM-refined texts to account for the diversity of LLMs and editing prompts.

\paragraph{Lack of Human Baseline} Similar to closely related studies on multi-author MGT detection \cite{zhang-etal-2024-llm,chakrabarty2024can,abassy2024llm}, our work does not present the human baseline results due to limited resources. We aim to establish the human baseline across various editing types in our future work.

%% file: sections/ethics.tex
\section*{Ethics Statement}
\paragraph{Expert-based Editing} Our team of expert editors is based in the United States and Canada, and their pay rate exceeds the corresponding hourly minimum wage. The annotation and voluntary survey results are collected and saved anonymously. The experts are warned about potentially sensitive and harmful content in the prompts and LLMs' responses related to various topics, including but not limited to politics, culture, sexual orientation, and religion. 

\paragraph{Use of AI-assistants} We use Grammarly\footnote{\href{https://app.grammarly.com}{\texttt{grammarly.com}}} to correct grammar, spelling, phrasing, and style errors in our paper. Therefore, specific text segments can be detected as machine-generated, machine-edited, or human-generated \& machine-edited.

\paragraph{Computational Costs} Evaluating an MGT detector on Beemo does not require any finetuning. To reduce the evaluation costs in \S\ref{sec:experimental_setup}, we pre-compute and save the detectors' predictions for each text version. This allows us to estimate the performance efficiently by manipulating these pre-computed results based on the task formulation. Further inference costs can be reduced with the help of distributed inference libraries (e.g., \texttt{accelerate}\footnote{\href{https://github.com/huggingface/accelerate}{\texttt{github.com/huggingface/accelerate}}} and \texttt{vllm}\footnote{\href{https://github.com/vllm-project/vllm}{\texttt{github.com/vllm-project/vllm}}}).

\paragraph{Potential Misuse} We acknowledge that Beemo can be misused for malicious purposes, including but not limited to training multi-author MGT detectors to evade further detection at scale. We release Beemo for research and development purposes and encourage responsible use of our benchmark.

\paragraph{Transparency} We release Beemo and all annotation materials following the standard open-source research practices. Our GitHub repository and HuggingFace dataset card provide comprehensive documentation on our benchmark creation process and data annotation guidelines.

\paragraph{Licensing Information} The prompts and human-written responses from \texttt{No Robots} are under the original dataset's license (CC-BY-NC-4.0). The MGTs and their LLM-edited versions are subject to the underlying instruction-finetuned LLMs' licensing terms (see \autoref{tab:lms}). The expert-edited MGTs are available under the MIT license, unless otherwise specified in the underlying instruction-finetuned LLMs' licensing terms.

%% file: sections/acknowledgement.tex
\section*{Acknowledgements}

We thank Nontobeko Magala for her contribution to the data collection, Alexey Artemov for his contribution to the figure design, Natalia Fedorova for her support, and our anonymous reviewers for their feedback.

%% file: sections/appendix_annotators.tex
\begin{table}[tp!]
\onecolumn
\section{Annotator Profiles}
\label{app:annotator_profiles}
    \vspace{0.1cm}
    \centering
    \begin{tabular}{lr}
    \toprule 
        \multicolumn{2}{c}{\textbf{What gender do you identify as?}}\\
         Male & 30.0\% \\
         Female & 70.0\% \\
         Nonbinary / other & 0\% \\ \midrule
         \multicolumn{2}{c}{\textbf{What ethnicity do you identify as?}}\\
         White / Caucasian & 58.8\% \\
         South Asian & 5.9\% \\
         Indigenous / Native American / Alaskan Native  & 0.0\% \\
         East Asian & 0.0\% \\
         Middle Eastern & 0.0\% \\
         Latinx & 0.0\% \\
         Black / African & 17.6\% \\
         Mixed / Mixed race & 11.8\% \\
         \midrule 

         \multicolumn{2}{c}{\textbf{What is your nationality?}}\\
         American & 17.5\% \\
         British & 17.5\% \\
         Indian & 5.9\% \\
         South African &  17.5\% \\
         Canadian & 17.5\% \\
         Kenyan &  11.8\% \\
         Malaysian & 5.9\% \\
         Irish  & 5.9\% \\ \midrule
         \multicolumn{2}{c}{\textbf{What is your native language?}}\\
         English & 75\% \\
         Multiple languages including English & 25\% 
         \\ \midrule
         \multicolumn{2}{c}{\textbf{What is your age?}}\\
         20-29 & 53.2\% \\
         30-39 & 17.5\% \\
         40-49 & 17.5\% \\
         50-59 & 11.8\% \\
         60+ & 0\% \\ \midrule
         \multicolumn{2}{c}{\textbf{What is your highest attained level of education?}}\\
         High school degree & 0.0\% \\
         Undergraduate degree & 52.9\% \\
         Postgraduate degree & 23.5\% \\
         Master's degree & 23.5\% \\
         Doctorate degree & 0\% \\ \midrule
        \multicolumn{2}{c}{\textbf{How many years of work experience do you have?}}\\
         1-3 years & 17.6\% \\
         4-6 years & 11.8\% \\ 
         7-9 years & 17.6\% \\ 
         10-12 years & 17.6\% \\ 
         13-15 years & 11.8\% \\ 
         16+ years & 17.6\% \\ 
         \bottomrule
    \end{tabular}
    \caption{Annotator profiles. Voluntary survey results from 17 respondents.}
    \label{tab:labeler_demographics}
\end{table}

%% file: sections/appendix_guidelines.tex
\begin{table}[tp!]
\onecolumn
\section{Annotation guidelines}
\label{section:appendix_guidelines}

    \centering
    \begin{minipage}[t]{0.95\linewidth}
    \par\noindent\rule{\textwidth}{1pt}

    \textbf{Objective}
    \vspace{0.1cm}

    The goal of this task is to refine machine-generated texts to make them more \textbf{human-like} in terms of fluency, factual correctness, coherence, and style.
    \vspace{0.2cm}

    \textbf{Task}
    \vspace{0.1cm}
    
    You are provided with an instruction pair, comprising two components: (1) an instruction given to an AI assistant, such as ChatGPT, and (2) the assistant’s response. Your objective is to refine the response, making it sound more human-like and ensuring it is free of errors.

    \vspace{0.2cm}

    \textbf{How many edits should be done?}
    \vspace{0.1cm}

    Attempt to edit 20\% to 40\% of the provided text. Reject the text if it requires more significant improvements or does not follow the instructions completely. Aim for real-life settings: if a text misses the mark or requires a lot of refinement, most likely, the user would ask the AI assistant for a better response.

    \vspace{0.1cm}

    To \textbf{reject} the text, hit the ``\textbf{Too bad}'' button. If there is \textbf{nothing to edit} in the text, hit the ``\textbf{Already perfect}'' button and proceed to the next example. No need to fill in the ``\textbf{Edit the AI Response}'' field in either of these two cases.
    
    \begin{enumerate}[noitemsep,topsep=0.1pt,leftmargin=1.2em]
        \item \textbf{Read:} Carefully read the provided text, paying attention to the instruction and to how well the AI assistant followed it.
        
        \item \textbf{Check:} Check that the prompt matches the category. See the fields \textbf{Category} and \textbf{Prompt} at the top part of the interface. If the prompt doesn’t fall into the specified category, reject the text by hitting the ``\textbf{Too bad}'' button.
        
        \item \textbf{Assess:} Evaluate the AI assistant's response for factual correctness, coherence, grammar, style, and overall human-like quality. Identify any errors or areas that need improvement.
        
        \item \textbf{Edit:} Make necessary revisions to the AI assistant's response to enhance its human-like quality. This may include rephrasing sentences, correcting grammar mistakes, ensuring coherence, and adding a personal touch to it. See the list of the issues you may want to correct below. We provide the recommended range of edits 20\%--40\% but it is not strict; make the edits based on your experience in editing and working with data generated by language models.

        \item \textbf{Proofread:} Review the edited response to ensure it is free of errors and flows smoothly. Check for any remaining issues or inconsistencies.

        \item \textbf{Submit:} Once you are satisfied with the edits and the AI response meets the criteria of being human-like and error-free, submit your final version.
    \end{enumerate}
\par\noindent\rule{\textwidth}{1pt}
\end{minipage}
\caption{A shortened version of the annotation guidelines for expert-based editing in \S\ref{subsec:expert_editing} (Part 1). The full version with the editing examples are provided in our GitHub repository at \href{https://github.com/Toloka/beemo/tree/main/guidelines}{\texttt{github.com/Toloka/beemo/guidelines}}.}
\end{table}

\begin{table}[tp!]
    \centering
    \begin{minipage}[t]{0.99\linewidth}
    \par\noindent\rule{\textwidth}{1pt}
    \subsection*{Issues of AI-generated text that you may want to correct}

    \emph{This list is not exhaustive. Follow your intuition and edit the text where you feel it's reasonable to do so.}

    \begin{enumerate}[noitemsep,topsep=0.1pt,leftmargin=1.2em]
        \item \textbf{Repetitions.} The same word, phrase, or sentence is repeated multiple times.
        \vspace{0.03cm}
        
        \textbf{Do:} Remove repeated words, phrases, and sentences. \colorbox{Gray}{This is an editing example.}

        \item \textbf{Awkward phrasing.} The text contains awkward or unnatural phrasing that makes it difficult to read or understand. It contains generic and simplistic language or, on the contrary, uses unnecessarily sophisticated and outdated words and phrases. 
        \vspace{0.03cm}
        
        \textbf{Do:} Improve the response for natural flow. \colorbox{Gray}{This is an editing example.}

        \item \textbf{Tone and style.} The tone and style of the text do not align with the intended audience and purpose. \vspace{0.03cm}
        
        \textbf{Do:} Adjust the tone and style. \colorbox{Gray}{This is an editing example.}

        \item \textbf{Texts written from the AI assistant's point of view.} AI assistants tend to add some unnecessary sentences to the beginning or the end of the text. These AI’s introductory sentences are unnecessary and should be omitted.
        \vspace{0.03cm}
        
        \textbf{Do:} Remove markers of a text generated by an AI assistant. \colorbox{Gray}{This is an editing example.}

        \item \textbf{Grammar errors.} The text is ungrammatical and contains unacceptable sentences.
        \vspace{0.03cm}
        
        \textbf{Do:} Edit so the text is grammatically correct, fix the sentence structure and correct the typos. \colorbox{Gray}{This is an editing example.}

        \item \textbf{Relevance.} The response does not fully follow the provided instruction.
        \vspace{0.03cm}
        
        \textbf{Do:} Remove any irrelevant or unnecessary information from the AI response. Add important details or context. If the AI response is completely irrelevant to the instruction, reject the text. \colorbox{Gray}{This is an editing example.} 
        
        \item \textbf{Personal touch.} The text is too generic.
        \vspace{0.03cm}
        
        \textbf{Do:} You may want to add a personal touch to make the text more engaging.  \colorbox{Gray}{This is an editing example.}

        \item \textbf{Fact-check the AI response.} Check the facts provided by the AI assistant. If you can fix the facts based on common knowledge or your personal experience, you don’t need to provide anything. If you searched online for the information to fix the facts, provide links to the found sources in the ``\textbf{Comment}'' that you used to fact-check the AI response.
        \vspace{0.03cm}
        
        \textbf{Do:} Decide to what degree it is possible to edit the AI response to improve factual correctness. If it requires substantial changes, discard the text. \colorbox{Gray}{This is an editing example.} 

        \item \textbf{Incomplete response.} If the response is interrupted accidentally, complete it. \colorbox{Gray}{This is an editing example.} 
    
        \item \textbf{Other.} You may find other issues that you want to edit. Let us know about them in ``\textbf{Comment}''.
    \end{enumerate}
    
\par\noindent\rule{\textwidth}{0.5pt}    
\underline{Example of web interface}
\vspace{0.02cm}

\textbf{Category}
\vspace{0.05cm}
\colorbox{Gray}{This is a toy category.}

\textbf{Prompt}
\vspace{0.05cm}
\colorbox{Gray}{This is a toy prompt.}

\textbf{AI Response}
\vspace{0.05cm}
\colorbox{Gray}{This is a toy LLM's response.}

\begin{itemize}[noitemsep,topsep=0pt]
    \item[\radiobutton] Too bad 
    \item[\radiobutton] Already perfect
\end{itemize}

\textbf{Edit the AI Response}

\fbox{
    \begin{minipage}{0.4\textwidth}
    \parbox{0.4\textwidth}{
        \centering
        \tiny
     }
     \end{minipage}
}

\textbf{Comment}

\fbox{
    \begin{minipage}{0.4\textwidth}
    \parbox{0.4\textwidth}{
        \centering
        \tiny
     }
     \end{minipage}
}

Proofread before submission. Thank you!

\par\noindent\rule{\textwidth}{1pt}
 
    \end{minipage}
    \caption{A shortened version of the annotation guidelines and an example of web interface for expert-based editing in \S\ref{subsec:expert_editing} (Part 2). The full version with the editing examples are provided in our GitHub repository at \href{https://github.com/Toloka/beemo/tree/main/guidelines}{\texttt{github.com/Toloka/beemo/guidelines}}.}
\end{table}

%% file: sections/appendix_issues.tex
\begin{table}[htp!]
\section{Examples of Common Issues in Machine-generated Texts} \label{appendix:issues}
\centering
\scriptsize
\begin{tabular}{p{0.2\textwidth} p{0.2\textwidth} p{0.23\textwidth} p{0.23\textwidth}}
\toprule
\textbf{Prompt} & \textbf{Model Output} & \textbf{Human Edit/Correct Answers} & \textbf{Explanation} \\
\midrule
\textbf{Format Inconsistencies} & 
\textit{\textbf{Category}: Generation (Llama2-7B-chat-hf)}\\
Write a shopping list for everything you need to make burritos. & 
Sure! Here are five essential items you'll need for burritos:  
\begin{enumerate}[nosep]
    \item \textbf{Tortillas}  
    \item \textbf{Meat}  
    \item \textbf{Vegetables}  
    \item \textbf{Cheese}  
    \item \textbf{Sauce}
\end{enumerate} & 
\begin{enumerate}[nosep]
    \item Flour tortillas  
    \item Cooked seasoned ground beef or shredded chicken  
    \item Black beans or refried beans  
    \item Chopped lettuce, tomatoes, and onions  
    \item Shredded cheese (cheddar or Mexican blend)  
    \item Sour cream or guacamole  
    \item Salsa or hot sauce  
    \item Rice (optional)
\end{enumerate} & 
The model-generated list is \textbf{too vague and incomplete}, making it \textbf{unhelpful for someone trying to shop for burrito ingredients}. The human edit \textbf{expands and specifies} the required items, ensuring \textbf{completeness and clarity}. \\[1em]
\midrule
\textbf{Hallucinations and Factual Errors} & 
\textit{\textbf{Category}: Open QA (Tulu-2-13B)}\\
What are the colors that light up in the children's game Simon? & 
Simon is a children's game that involves four colors: red, blue, \textbf{orange}, and green. & 
Simon is a popular electronic game that features four colored buttons: red, blue, \textbf{yellow}, and green. Players must follow an increasing sequence of lights and sounds by pressing the correct colors in order. & 
The model provides an incorrect answer by listing \textbf{orange} instead of \textbf{yellow} as one of the colors in the game Simon. The human edit corrects the factual inaccuracy while also providing additional context about the game. \\[1em]
\midrule
\textbf{Struggling with Creative Writing} & 
\textit{\textbf{Category}: Generation (Llama2-7B-chat-hf)}\\
Write a short poem about a sheep adventuring on a hill. & 
Sure! Here is a short poem about a sheep adventuring on a hill:
The sheep goes up, the sheep goes down,  
Wandering freely all around.  
Soft white wool and tiny feet,  
Through the fields it loves to greet. & 
A lone sheep climbs the emerald crest,  
Chasing dreams with fearless zest.  
Soft wool dancing in the breeze,  
Leaping high with nimble ease. & 
This example highlights how the model struggles with poetic creativity, often defaulting to simplistic, predictable rhymes. The human edit improves imagery, rhythm, and poetic depth. \\[1em]
\midrule
\textbf{Complicated and Repetitive Vocabulary} & 
\textit{\textbf{Category}: Summarize (Mistral-7B-Instruct)}\\
Give me a TL;DR of what Professor Kaptchuk is researching. & 
Professor Ted Kaptchuk of Harvard-affiliated Beth Israel Deaconess Medical Center is researching the placebo effect. His work explores how the brain and body react to placebos, and he is particularly interested in understanding the biological, neurological, and psychological mechanisms underlying placebo responses. He studies how factors such as doctor-patient interactions, belief systems, and environmental cues influence placebo responses. His research provides insights into how placebos can be integrated into medical treatments to enhance therapeutic outcomes and improve patient care. & 
The placebo effect goes beyond positive thinking. Professor Ted Kaptchuk researches how doctor-patient interactions and environmental factors contribute to the body's response to placebos, influencing medical treatment outcomes. & 
The model output is overly verbose and contains repetitive phrasing about the placebo effect. The human edit distills the key information concisely while retaining clarity and relevance. \\[1em]
\midrule
\textbf{Lack of Natural Flow and Repetitive Structure} & 
\textit{\textbf{Category}: Rewrite (Gemma-2B-IT)}\\
Please rewrite this as a short love note one would write on a post-it. & 
My dearest S,  
Focus. You and you. It's the little moments that make life special. You are the sunshine that brightens my day, the melody that lifts my soul, the laughter that warms my heart. You are everything wonderful, and I am grateful for you every single day. Always yours. & 
Us, U, and S are the last two letters of FOCUS. That's what you are to me—always in focus. Love you. & 
The model output is \textbf{overly long and repetitive}, making it sound unnatural for a simple post-it note. The human edit keeps it \textbf{concise, poetic, and engaging}, aligning with the task's request for a short love note. \\
\bottomrule
\end{tabular}
\caption{Examples of common issues in MGTs along with corresponding human edits. The examples are on the issues described in \S\ref{subsec:expert_editing}: (i) format inconsistencies; (ii) hallucinations and factual errors; (iii) challenges in creative writing; (iv) overly complicated and repetitive vocabulary; and (v) lack of natural flow with repetitive structure.}
\label{tab:common_issues_examples}
\end{table}

%% file: sections/appendix_LLM_edits.tex
\begin{table*}[htbp]
\section{Details on LLM-based Editing}
\label{appendix:llm_editing}
\setlength{\tabcolsep}{4pt}
\centering
\scriptsize
\begin{tabularx}{\textwidth}{@{}c>{\raggedright\arraybackslash}X>{\raggedright\arraybackslash}X>{\raggedright\arraybackslash}X>{\raggedright\arraybackslash}X>{\raggedright\arraybackslash}X@{}}
\toprule
 & \textbf{Prompt} & \textbf{\gemmas} & \textbf{Expert-edited} & \textbf{\llamaxl-edited} & \textbf{\gpt-edited} \\
\midrule
P1 & You are given a prompt and a text generated by AI using this prompt. \uline{Your task is to edit the AI-generated text to make it sound human-like and error-free}. Ensure your overall edits do not exceed 40\% of the generated text and the edited text follows the user request. Output only the edited text and do not explain your edits. \begin{promptbox}
Prompt: \{prompt\}
\end{promptbox}
\begin{llmtextbox}
AI text: \{model\_output\}
\end{llmtextbox} & According to the statement, Chick-fil-A's core values are purpose, culture, and teamwork. They believe that through these values, they can provide a positive work experience for their staff and ensure that they deliver the best possible service to their customers. & \unchangedbox{According} \unchangedbox{to} \changedbox{this} \unchangedbox{statement,} \unchangedbox{Chick-fil-A} \changedbox{does} \changedbox{its} \changedbox{best} \changedbox{work} \changedbox{through} \unchangedbox{teamwork} \changedbox{and} \changedbox{collaboration.} & \unchangedbox{According} \unchangedbox{to} \unchangedbox{the} \unchangedbox{statement,} \unchangedbox{Chick-fil-A's} \unchangedbox{core} \unchangedbox{values} \changedbox{reflect} \unchangedbox{their} \unchangedbox{purpose,} \unchangedbox{culture,} \changedbox{and} \changedbox{team} \changedbox{experience.} \changedbox{They} \changedbox{believe} \changedbox{that} \changedbox{by} \changedbox{putting} \changedbox{the} \changedbox{needs} \changedbox{of} \changedbox{their} \changedbox{Operators,} \changedbox{Team} \changedbox{Members,} \changedbox{and} \changedbox{customers} \changedbox{first,} \changedbox{they} \changedbox{can} \changedbox{do} \changedbox{their} \changedbox{best} \changedbox{work,} \changedbox{particularly} \changedbox{when} \changedbox{achieved} \changedbox{through} \unchangedbox{teamwork} \changedbox{and} \changedbox{collaboration.} & \unchangedbox{According} \unchangedbox{to} \unchangedbox{the} \unchangedbox{statement,} \unchangedbox{Chick-fil-A} \changedbox{believes} \changedbox{they} \changedbox{do} \changedbox{their} \changedbox{best} \changedbox{work} \changedbox{through} \unchangedbox{teamwork} \changedbox{and} \changedbox{collaboration.} \changedbox{By} \changedbox{fostering} \changedbox{an} \changedbox{inclusive} \changedbox{culture} \changedbox{and} \changedbox{leveraging} \changedbox{the} \changedbox{strengths} \changedbox{of} \changedbox{their} \changedbox{diverse} \changedbox{talent,} \changedbox{they} \changedbox{aim} \changedbox{to} \changedbox{serve} \changedbox{their} \changedbox{Operators,} \changedbox{Team} \changedbox{Members,} \changedbox{and} \changedbox{customers} \changedbox{effectively.} \\
\midrule
P2 & You are given a pair containing two components: (1) a user prompt for an AI assistant and (2) the AI assistant's response. \uline{Refine the AI-generated response to make it sound more natural}. Vary your editing patterns and the portions of text you choose to modify, and ensure your overall edits are 20-40\% of the words in the response. \begin{promptbox}
User prompt: \{prompt\}
\end{promptbox}
\begin{llmtextbox}
AI-generated response: \{model\_output\}
\end{llmtextbox} & According to the statement, Chick-fil-A's core values are purpose, culture, and teamwork. They believe that through these values, they can provide a positive work experience for their staff and ensure that they deliver the best possible service to their customers. & \unchangedbox{According} \unchangedbox{to} \changedbox{this} \unchangedbox{statement,} \unchangedbox{Chick-fil-A} \changedbox{does} \changedbox{its} \changedbox{best} \changedbox{work} \changedbox{through} \unchangedbox{teamwork} \changedbox{and} \changedbox{collaboration.} & \unchangedbox{According} \unchangedbox{to} \unchangedbox{Chick-fil-A's} \unchangedbox{statement,} \changedbox{they} \changedbox{attribute} \changedbox{their} \changedbox{best} \changedbox{work} \changedbox{to} \changedbox{a} \changedbox{combination} \changedbox{of} \changedbox{their} \unchangedbox{core} \unchangedbox{values,} \changedbox{which} \changedbox{emphasize} \changedbox{the} \changedbox{importance} \changedbox{of} \unchangedbox{teamwork,} \changedbox{a} \changedbox{strong} \changedbox{company} \unchangedbox{culture,} \changedbox{and} \changedbox{staying} \changedbox{true} \changedbox{to} \changedbox{their} \unchangedbox{purpose.} \changedbox{By} \changedbox{prioritizing} \changedbox{these} \changedbox{values,} \changedbox{they} \changedbox{aim} \changedbox{to} \changedbox{foster} \changedbox{a} \changedbox{positive} \changedbox{work} \changedbox{environment} \changedbox{for} \changedbox{their} \changedbox{staff} \changedbox{while} \changedbox{delivering} \changedbox{exceptional} \changedbox{service} \changedbox{to} \changedbox{their} \changedbox{customers.} & \unchangedbox{According} \unchangedbox{to} \unchangedbox{the} \unchangedbox{statement,} \unchangedbox{Chick-fil-A's} \changedbox{best} \changedbox{work} \changedbox{is} \changedbox{guided} \changedbox{by} \changedbox{their} \unchangedbox{core} \unchangedbox{values} \changedbox{of} \unchangedbox{purpose,} \unchangedbox{culture,} \changedbox{and} \unchangedbox{teamwork.} \changedbox{By} \changedbox{embracing} \changedbox{these} \changedbox{values,} \changedbox{they} \changedbox{aim} \changedbox{to} \changedbox{create} \changedbox{a} \changedbox{positive} \changedbox{work} \changedbox{environment} \changedbox{for} \changedbox{their} \changedbox{staff} \changedbox{and} \changedbox{deliver} \changedbox{exceptional} \changedbox{service} \changedbox{to} \changedbox{their} \changedbox{customers.} \\
\midrule
P3 & Modify a machine-generated response to a given prompt to \uline{make it appear more like it was written by a native English speaker}. Ensure the revised version follows the user's intent. You should just give me the revised version without any other words. \begin{promptbox}
Prompt: \{prompt\}
\end{promptbox}
\begin{llmtextbox}
Machine-generated response: \{model\_output\}
\end{llmtextbox} & According to the statement, Chick-fil-A's core values are purpose, culture, and teamwork. They believe that through these values, they can provide a positive work experience for their staff and ensure that they deliver the best possible service to their customers. & \unchangedbox{According} \unchangedbox{to} \changedbox{this} \unchangedbox{statement,} \unchangedbox{Chick-fil-A} \changedbox{does} \changedbox{its} \changedbox{best} \changedbox{work} \changedbox{through} \unchangedbox{teamwork} \changedbox{and} \changedbox{collaboration.} & \unchangedbox{According} \unchangedbox{to} \unchangedbox{Chick-fil-A's} \unchangedbox{statement,} \changedbox{they} \changedbox{do} \changedbox{their} \changedbox{best} \changedbox{work} \changedbox{when} \changedbox{they} \changedbox{operate} \changedbox{as} \changedbox{a} \changedbox{team,} \changedbox{leveraging} \changedbox{the} \changedbox{strengths} \changedbox{of} \changedbox{their} \changedbox{diverse} \changedbox{talent} \changedbox{through} \unchangedbox{collaboration} \changedbox{and} \changedbox{inclusivity.} & \unchangedbox{According} \unchangedbox{to} \unchangedbox{the} \unchangedbox{statement,} \unchangedbox{Chick-fil-A} \changedbox{believes} \changedbox{they} \changedbox{do} \changedbox{their} \changedbox{best} \changedbox{work} \changedbox{through} \changedbox{their} \unchangedbox{core} \changedbox{values} \changedbox{of} \changedbox{service,} \unchangedbox{teamwork,} \changedbox{purpose-driven} \changedbox{actions,} \changedbox{and} \changedbox{continuous} \changedbox{improvement.} \changedbox{These} \changedbox{values} \changedbox{help} \changedbox{create} \changedbox{a} \changedbox{positive} \changedbox{work} \changedbox{experience} \changedbox{for} \changedbox{their} \changedbox{staff} \changedbox{and} \changedbox{ensure} \changedbox{they} \changedbox{deliver} \changedbox{the} \changedbox{best} \changedbox{possible} \changedbox{service} \changedbox{to} \changedbox{their} \changedbox{customers.} \\
\bottomrule
\end{tabularx}
\caption{An example of edits made by the experts, \llamaxl, and \gpt for a \gemmas's response on Chick-fil-A Values Statement. \unchangedbox{Blue boxes} indicate individual words that remain unchanged from the original Model Output. \changedbox{Pink boxes} indicate individual words that were changed or added in the edited versions.}
\label{tab:chick-fil-a-values-comparison}
\end{table*}

%% file: sections/appendix_analysis.tex
\onecolumn

\section{Details on Benchmark Analysis}
\label{appendix:analysis}
\input{tables/stats_model}

\input{tables/bertscore}


\begin{figure}[h!]
    \centering
    \includegraphics[width=0.7\linewidth]{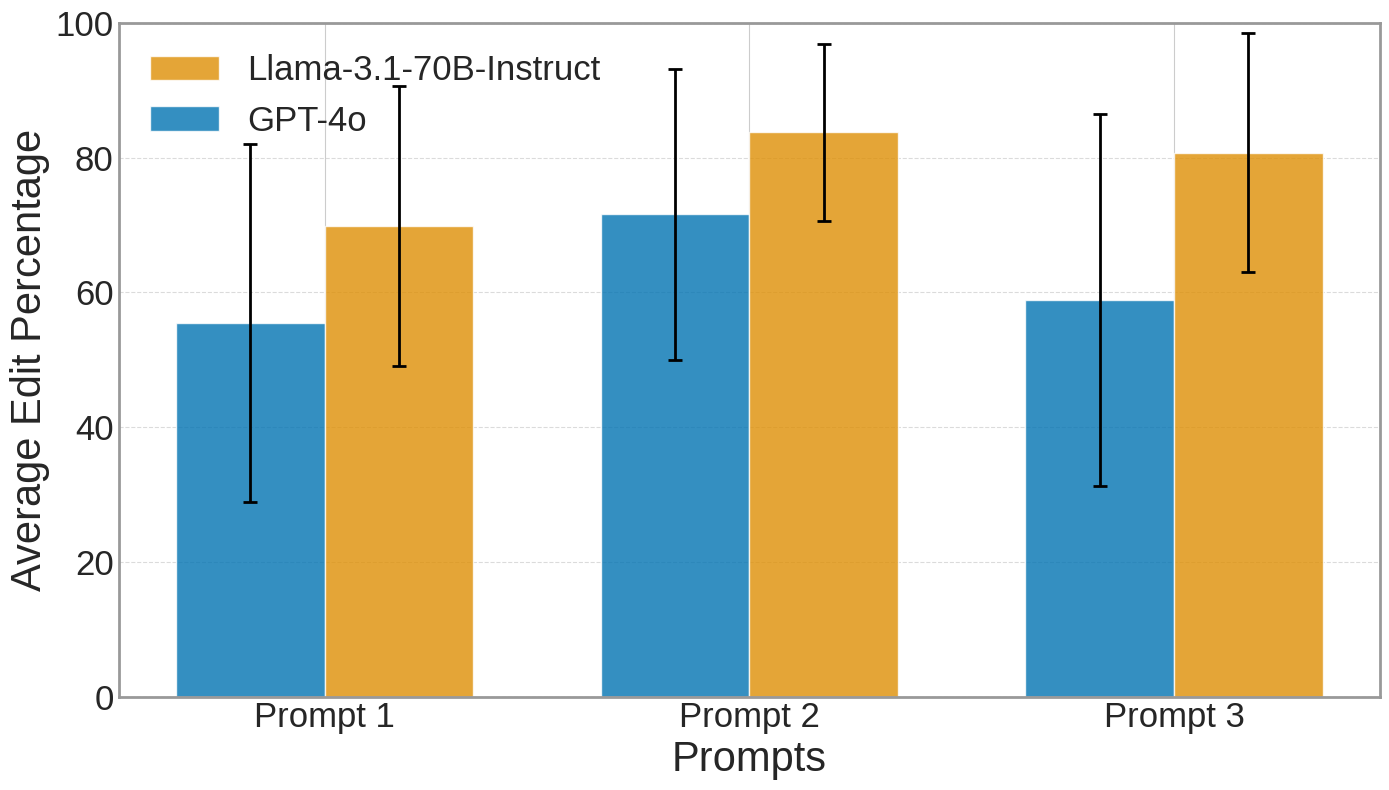}
    \caption{Comparison of average edit percentages between \gpt and \llamaxl across three different prompts. The bars represent the mean edit percentage for each prompt, with error bars indicating the standard deviation.}
    \label{fig:prompt-comparison}
\end{figure}

\input{tables/survey}

%% file: tables/stats_model.tex
\begin{table*}[htp!]
    \centering
     \resizebox{\textwidth}{!}{
    \begin{tabular}{lrrrrrrrr}
\toprule
\textbf{Model} & \textbf{\# Examples} & \textbf{\# Tokens} & \textbf{\# Stopwords} & \textbf{FKG} & \textbf{LD(M, H)} & \textbf{LD(M, E)} & \textbf{LD(M, L)} & \textbf{LD(M, G)} \\
\midrule
\zephyr & 229 & 203.99 \small $\pm$ 143.25 & 98.43 \small $\pm$ 76.85 & 9.41 \small $\pm$ 6.27 & 211.69 \small $\pm$ 151.6 & 144.93 \small $\pm$ 111.01 & 189.86 \small $\pm$ 151.39 & 160.77 \small $\pm$ 151.02 \\
\tulus & 215 & 140.35 \small $\pm$ 95.03 & 70.9 \small $\pm$ 50.5 & 8.42 \small $\pm$ 3.64 & 164.92 \small $\pm$ 126.19 & 104.53 \small $\pm$ 95.91 & 127.43 \small $\pm$ 91.15 & 97.92 \small $\pm$ 92.36 \\
\tulum & 211 & 149.95 \small $\pm$ 113.37 & 74.15 \small $\pm$ 60.96 & 9.44 \small $\pm$ 7.08 & 170.25 \small $\pm$ 142.93 & 92.55 \small $\pm$ 93.53 & 136.29 \small $\pm$ 112.99 & 103.84 \small $\pm$ 111.87 \\
\gemmas & 204 & 153.79 \small $\pm$ 137.75 & 69.59 \small $\pm$ 65.49 & 7.73 \small $\pm$ 3.65 & 188.56 \small $\pm$ 172.0 & 120.98 \small $\pm$ 130.15 & 157.91 \small $\pm$ 141.57 & 129.9 \small $\pm$ 139.24 \\
\gemmam & 209 & 140.81 \small $\pm$ 111.14 & 63.71 \small $\pm$ 52.87 & 8.01 \small $\pm$ 4.06 & 175.59 \small $\pm$ 145.4 & 101.6 \small $\pm$ 106.03 & 143.42 \small $\pm$ 114.69 & 113.28 \small $\pm$ 110.16 \\
\llamas & 234 & 229.15 \small $\pm$ 154.72 & 107.67 \small $\pm$ 81.07 & 8.63 \small $\pm$ 3.70 & 241.4 \small $\pm$ 170.31 & 160.38 \small $\pm$ 132.92 & 212.13 \small $\pm$ 163.73 & 183.9 \small $\pm$ 164.31 \\
\llamam & 234 & 212.42 \small $\pm$ 150.89 & 102.86 \small $\pm$ 77.34 & 7.77 \small $\pm$ 3.44 & 217.61 \small $\pm$ 163.6 & 166.83 \small $\pm$ 131.14 & 194.08 \small $\pm$ 155.4 & 164.98 \small $\pm$ 153.9 \\
\llamal & 224 & 239.20 \small $\pm$ 146.80 & 118.68 \small $\pm$ 81.02 & 7.98 \small $\pm$ 4.56 & 256.59 \small $\pm$ 157.17 & 190.53 \small $\pm$ 134.77 & 226.63 \small $\pm$ 154.74 & 198.96 \small $\pm$ 154.56 \\
\mistral & 201 & 146.07 \small $\pm$ 120.86 & 72.48 \small $\pm$ 64.95 & 8.93 \small $\pm$ 7.45 & 154.20 \small $\pm$ 128.54 & 101.72 \small $\pm$ 102.59 & 134.58 \small $\pm$ 120.5 & 105.77 \small $\pm$ 120.43 \\
\mixtral & 226 & 192.91 \small $\pm$ 137.39 & 92.37 \small $\pm$ 73.09 & 8.56 \small $\pm$ 3.83 & 204.11 \small $\pm$ 154.21 & 117.62 \small $\pm$ 99.72 & 174.14 \small $\pm$ 142.85 & 140.6 \small $\pm$ 143.09 \\
\vspace{-1em}\\ \cdashline{1-9} \vspace{-1em}\\
\texttt{Overall} & 2,187 & 182.53 \small $\pm$ 137.88 & 87.93 \small $\pm$ 71.87 & 8.49 \small $\pm$ 4.98 & 199.43 \small $\pm$ 155.3 & 131.35 \small $\pm$ 119.36 & 171.08 \small $\pm$ 141.29 & 141.43 \small $\pm$ 140.77 \\
\bottomrule
\end{tabular}
}
    \caption{General statistics by LLM. \textbf{FKG}=Flesch-Kincaid Grade; \textbf{H}=human-written; \textbf{M}=machine-generated; \textbf{E}=expert-edited; \textbf{L}=\llamaxl-edited; \textbf{G}=\gpt-edited;  \textbf{FKG}=Flesch-Kincaid Grade; \textbf{LD(M, H/E/L/G)}=Levenshtein distance. \textbf{L} and \textbf{G} are aggregated over three prompts (\S\ref{subsec:llm_editing}).}
    \label{tab:stats_model}
\end{table*}

%% file: tables/bertscore.tex
\begin{table*}[ht!]
\centering
\resizebox{\textwidth}{!}{%
\begin{tabular}{lrrrr}
\toprule
& \textbf{BERTScore(M,H)} & \textbf{BERTScore(M,E)} & \textbf{BERTScore(M,L)} & \textbf{BERTScore(M,G)}  \\
\midrule
\zephyr  & 0.72 \small $\pm$ 0.05 & 0.83 \small $\pm$ 0.07 & 0.79 \small $\pm$ 0.06 & 0.84 \small $\pm$ 0.07 \\
\tulus   & 0.73 \small $\pm$ 0.06 & 0.85 \small $\pm$ 0.07 & 0.79 \small $\pm$ 0.06 & 0.86 \small $\pm$ 0.07 \\
\tulum   & 0.73 \small $\pm$ 0.06 & 0.83 \small $\pm$ 0.08 & 0.79 \small $\pm$ 0.06 & 0.85 \small $\pm$ 0.07 \\
\gemmas  & 0.71 \small $\pm$ 0.06 & 0.81 \small $\pm$ 0.09 & 0.77 \small $\pm$ 0.06 & 0.83 \small $\pm$ 0.07 \\
\gemmam  & 0.72 \small $\pm$ 0.06 & 0.83 \small $\pm$ 0.09 & 0.78 \small $\pm$ 0.06 & 0.84 \small $\pm$ 0.06 \\
\llamas  & 0.73 \small $\pm$ 0.06 & 0.80 \small $\pm$ 0.07 & 0.79 \small $\pm$ 0.05 & 0.84 \small $\pm$ 0.07 \\
\llamam  & 0.71 \small $\pm$ 0.06 & 0.80 \small $\pm$ 0.09 & 0.76 \small $\pm$ 0.07 & 0.81 \small $\pm$ 0.08 \\
\llamal  & 0.72 \small $\pm$ 0.06 & 0.83 \small $\pm$ 0.07 & 0.79 \small $\pm$ 0.06 & 0.83 \small $\pm$ 0.07 \\
\mistral & 0.74 \small $\pm$ 0.06 & 0.83 \small $\pm$ 0.08 & 0.79 \small $\pm$ 0.06 & 0.85 \small $\pm$ 0.07 \\
\mixtral & 0.74 \small $\pm$ 0.05 & 0.85 \small $\pm$ 0.08 & 0.80 \small $\pm$ 0.06 & 0.86 \small $\pm$ 0.07 \\
\cdashline{1-5}
\texttt{Overall} & 0.72 \small $\pm$ 0.06 & 0.83 \small $\pm$ 0.08 & 0.79 \small $\pm$ 0.06 & 0.84 \small $\pm$ 0.07 \\
\bottomrule
\end{tabular}
}
\caption{Similarity scores by LLM. \textbf{H} = human-written, \textbf{M} = machine-generated, \textbf{E} = expert-edited; \textbf{L}=\llamaxl-edited; \textbf{G}=\gpt-edited; \textbf{BERTScore(M, H/E/L/G)}=BERTScore similarity. \textbf{L} and \textbf{G} are aggregated over three prompts (\S\ref{subsec:llm_editing}).}
\label{tab:bertscore}
\end{table*}

%% file: tables/survey.tex
\begin{table}[tp!]

\section{Details on Annotation Survey}
\label{appendix:survey}

    \centering
    \begin{minipage}[t]{\linewidth}
    \par\noindent\rule{\textwidth}{1pt}

    \textbf{Survey: Editing AI-generates Responses}
    \vspace{0.1cm}

    Answer the following questions related to editing AI-generated responses. The questions are divided into two groups. First, you are asked to judge general issues in the AI-generated responses based on how often you have observed them. Second, you are asked to share your strategies in the editing AI-generated responses in the free form.
    \vspace{0.2cm}

    \emph{How often do you observe the following issues in the AI-generated responses?}
    \vspace{0.2cm}
    
    Rate each issue on a scale from 1 to 5, where 1 stands for ``never,'' 3 stands for ``sometimes,'' and 5 stands for ``very often.''
    
    \vspace{0.12cm}
    \textbf{Q1: Repetitions.} The same word, phrase, or sentence is repeated in the AI-generated text multiple times.

    \input{tables/likert}

    \vspace{0.12cm}
    \textbf{Q2: Awkward phrasing.} The AI-generated text contains awkward or unnatural phrasing that makes it difficult to read or understand. It contains generic and simplistic language or, on the contrary, uses unnecessarily sophisticated and outdated words and phrases.

\input{tables/likert}
    \vspace{0.12cm}
    \textbf{Q3: Tone and style.} The tone and style of the AI-generated text do not align with the user intent.

\input{tables/likert}
    \vspace{0.12cm}
    \textbf{Q4: Grammar errors.} The AI-generated text is ungrammatical and does not sound natural.

\input{tables/likert}
    \vspace{0.12cm}
    \textbf{Q5: Relevance.} The AI-generated text does not fully follow the user's prompt or complete the target task.

\input{tables/likert}
    \vspace{0.12cm}
    \textbf{Q6: Code fragments.} The AI-generated text includes fragments that resemble programming language and are not contextually relevant.

\input{tables/likert}
    \vspace{0.12cm}
    \textbf{Q7: Incomplete response.} The AI-generated text is incomplete.

\input{tables/likert}
    \vspace{0.12cm}
    \textbf{Q8: Factual inconsistency.} The facts described in the AI-generated texts are wrong.

\input{tables/likert}
    \vspace{0.12cm}
    \textbf{Q9: Incoherence.} Segments of the AI-generated text are not connected or contradict each other.

\input{tables/likert}
    \par\noindent\rule{\textwidth}{1pt}
    \end{minipage}
    \caption{A survey on AI-generated response issues and editing strategies (Part 1).}
    \label{tab:annotation_survey_p1}
\end{table}

\begin{table}[tp!]
    \centering
    \begin{minipage}[t]{\linewidth}
    \par\noindent\rule{\textwidth}{1pt}
    \vspace{.15cm}

    \emph{Answer the following questions in 2-5 sentences.}
    \vspace{0.1cm}

    1. Do you have specific strategies for editing AI-generated texts?

    \fbox{
        \begin{minipage}{0.5\textwidth}
        \parbox{0.5\textwidth}{
            \centering
            \tiny
        }
        \end{minipage}
    }
    
    \vspace{0.2cm}

    2. In your experience, how does editing AI-generated texts compare with editing texts written by humans?

    \fbox{
        \begin{minipage}{0.5\textwidth}
        \parbox{0.5\textwidth}{
            \centering
            \tiny
        }
        \end{minipage}
    }
    
    \vspace{0.2cm}

    3. What difficulties or challenges have you encountered when editing the AI-generated texts?

    \fbox{
        \begin{minipage}{0.5\textwidth}
        \parbox{0.5\textwidth}{
            \centering
            \tiny
        }
        \end{minipage}
    }
    
    \vspace{0.2cm}

    Thank you for your time and efforts. 
    
    \vspace{0.1cm}
    Leave a comment if you want to add something.
    
    \fbox{
        \begin{minipage}{0.5\textwidth}
        \parbox{0.5\textwidth}{
            \centering
            \tiny
        }
        \end{minipage}
    }

    \par\noindent\rule{\textwidth}{1pt}
    \end{minipage}
    \caption{A survey on AI-generated response issues and editing strategies (Part 2).}
    \label{tab:annotation_survey_p2}
    
    \vspace{1cm}
    

    \begin{minipage}[t]{\textwidth}
        \centering
        \includegraphics[width=0.9\textwidth]{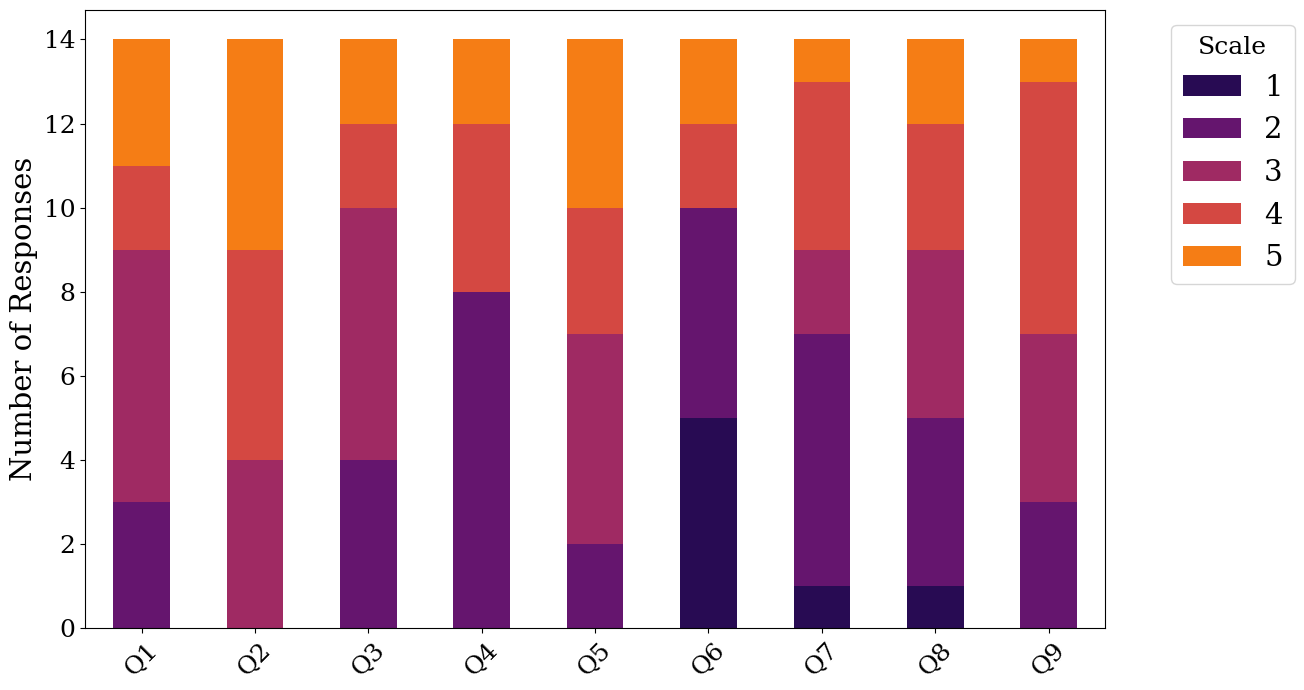}
    \captionof{figure}{Results of the survey on issues in the AI-generated responses. \textbf{Q1}=Repetitions; \textbf{Q2}=Awkward phrasing; \textbf{Q3}=Tone and style. \textbf{Q4}=Grammar errors; \textbf{Q5}=Relevance; \textbf{Q6}=Code fragments; \textbf{Q7}=Incomplete response; \textbf{Q8}=Factual inconsistency; \textbf{Q9}=Incoherence.}
    \label{fig1}
\end{minipage}
\end{table}

%% file: tables/likert.tex
\begin{center}
    \begin{tabular}{llccccccc}
    \centering
        & & 1 & 2 & 3 & 4 & 5 & \\
        & \textbf{Never} & \radiobutton & \radiobutton & \radiobutton & \radiobutton &  \radiobutton & \textbf{Very often} &  \\
    \end{tabular}
\end{center}

%% file: sections/appendix_setup.tex
\section{Details on MGT Detectors} \label{appendix:setup}
\subsection*{Zero-shot MGT Detectors}

\begin{itemize}
    \item Binoculars utilizes the ratio of perplexities of a text obtained using two pre-trained LLMs to compute cross-perplexity. This relative measure of perplexity from the lens of two LLMs helps detect LLM-generated texts. 
    \item Log Probability utilizes the average of the log probabilities of each token in a text and classifies text with a higher average log probability as machine-generated. This follows the notion that the probabilities of the tokens, as measured by the LLM that generated them, should be higher.
    \item Rank. In this approach, text with a higher average rank of each token in the text is determined to be machine-generated text. Rank is calculated by sorting the vocabulary tokens in decreasing order of likelihood.
    \item Log-Rank. This method classifies text with higher average values as machine-generated by calculating the average of the observed log-ranks of all tokens. 
    \item Entropy attributes higher entropy text as machine-generated. Entropy is calculated as the negative sum of the probability times log probabilities of all tokens at each position of the text. 
    \item DetectLLM has two variations: DetectLLM-LRR, which replaces log-likelihood computation with log-rank, and DetectLLM-NPR, which computes normalized perturbed log-rank. The former is fast and efficient, while the latter is slow but with higher detection accuracy.
    \item DetectGPT leverages the structure of LLMs' probability curve to identify machine-generated texts. Specifically, it assumes that LLMs' generations tend to occupy negative curvature regions.

\end{itemize}

\subsection*{Pretrained MGT Detectors}

\begin{itemize}
    \item RADAR uses an adversarial learning setup in which a paraphraser LM learns to alter the machine-generated text to evade detection. A detector LM then learns from this adversarially paraphrased text to better its performance.

    \item AIGC MPU augments LLM training with an additional length-sensitive multiscale positive-unlabeled (MPU) loss term and a text multiscaling module to enhance the training data. The method works with the assumption that MGT can be formulated as a partial Positive-Unlabeled (PU) problem by assuming short-length machine texts as partially unlabeled. 

     \item MAGE is the LongFormer model \cite{Beltagy2020Longformer} finetuned on 447K human-written and machine-generated texts collected from a wide range of sources.

     \item OpenAI's RoBERTa-based detectors: OpenAI released pretrained detectors using the RoBERTa models (base and large; \citealp{liu2019robertarobustlyoptimizedbert}) trained on text generated by GPT-2. 
\end{itemize}

%% file: sections/appendix_results.tex
\section{Details on Empirical Evaluation Results}
\label{appendix:results}
 
\input{tables/exp_results_by_category}

%% file: tables/exp_results_by_category.tex
\begin{table*}[ht!]
\centering
\setlength{\tabcolsep}{10pt}
{\renewcommand{\arraystretch}{1}}
\resizebox{\textwidth}{!}{
\begin{tabular}{llRRRRRRR}
\toprule
\multicolumn{2}{p{2ex}}{\textbf{Detector}} & \multicolumn{1}{c}{\textbf{H vs. M}} & \multicolumn{1}{c}{\textbf{E vs. M}} & \multicolumn{1}{c}{\textbf{H vs. E}} & \multicolumn{1}{c}{\textbf{L vs. M}} & \multicolumn{1}{c}{\textbf{H vs. L}} & \multicolumn{1}{c}{\textbf{G vs. M}} & \multicolumn{1}{c}{\textbf{H vs. G}} \\
\midrule

\multicolumn{9}{c}{\textbf{Zero-shot MGT Detectors}} \\ \midrule

\textbf{Binoculars} & & 93.64 & 82.84 & 76.19 & 75.26 & 81.80 & 74.44 & 77.03 \\ 
\midrule

\multirow{4}{*}{\textbf{Log Probability}} 
& \small{\texttt{\small{\texttt{GPT2-XL}}}} & 74.41 & 67.21 & 63.02 & 72.21 & 56.50 & 68.20 & 61.55 \\
& \small{\texttt{OPT-1.3B}} & 82.13 & 70.03 & 70.28 & 76.52 & 63.23 & 70.35 & 71.00 \\
& \small{\texttt{Falcon-7B}} & 90.44 & 71.53 & 83.59 & 77.07 & 80.20 & 71.01 & 84.83 \\
& \small{\texttt{Qwen2-7B}} & 91.14 & 75.85 & 81.57 & 58.62 & 87.95 & 67.23 & 84.76 \\
\midrule

\multirow{4}{*}{\textbf{Rank}} 
& \small{\texttt{GPT2-XL}} & 67.87 & 61.35 & 59.02 & 67.60 & 49.58 & 62.81 & 54.87 \\
& \small{\texttt{OPT-1.3B}} & 75.29 & 60.95 & 68.88 & 72.03 & 54.29 & 64.33 & 63.56 \\
& \small{\texttt{Falcon-7B}} & 83.47 & 62.21 & 78.61 & 72.42 & 65.54 & 64.37 & 74.84 \\
& \small{\texttt{Qwen2-7B}} & 81.74 & 64.23 & 73.14 & 54.74 & 78.87 & 59.49 & 76.01 \\
\midrule

\multirow{4}{*}{\textbf{Log-Rank}} 
& \small{\texttt{GPT2-XL}} & 76.28 & 68.08 & 63.65 & 74.53 & 54.43 & 69.61 & 61.05 \\
& \small{\texttt{OPT-1.3B}} & 80.64 & 69.97 & 68.07 & 77.28 & 58.28 & 71.04 & 67.01 \\
& \small{\texttt{Falcon-7B}} & 89.56 & 71.44 & 82.15 & 77.95 & 76.94 & 71.55 & 82.50 \\
& \small{\texttt{Qwen2-7B}} & 90.44 & 74.75 & 80.79 & 58.25 & 87.22 & 66.50 & 84.01 \\
\midrule

\multirow{4}{*}{\textbf{Entropy}} 
& \small{\texttt{GPT2-XL}} & 36.66 & 42.43 & 43.56 & 27.97 & 60.51 & 33.93 & 53.34 \\
& \small{\texttt{OPT-1.3B}} & 33.92 & 42.95 & 39.53 & 26.89 & 58.58 & 33.40 & 50.31 \\
& \small{\texttt{Falcon-7B}} & 19.50 & 37.62 & 23.40 & 25.86 & 38.92 & 30.69 & 32.09 \\
& \small{\texttt{Qwen2-7B}} & 8.26 & 26.53 & 14.41 & 42.18 & 10.31 & 34.35 & 12.36 \\
\midrule

\multirow{4}{*}{\makecell[l]{\textbf{DetectLLM} \\ \textbf{Likelihood Log-Rank Ratio}}} 
& \small{\texttt{GPT2-XL}} & 77.86 & 69.39 & 63.26 & 78.35 & 47.81 & 71.95 & 57.65 \\
& \small{\texttt{OPT-1.3B}} & 71.68 & 67.46 & 57.28 & 77.28 & 40.81 & 71.08 & 50.44 \\
& \small{\texttt{Falcon-7B}} & 83.28 & 68.72 & 73.11 & 78.37 & 60.35 & 71.06 & 68.87 \\
& \small{\texttt{Qwen2-7B}} & 84.26 & 68.21 & 73.92 & 56.07 & 80.81 & 62.14 & 77.37 \\
\midrule

\multirow{4}{*}{\makecell[l]{\textbf{DetectLLM} \\ \textbf{Normalized Perturbed Log-Rank}}} 
& \small{\texttt{GPT2-XL}} & 71.76 & 66.26 & 69.49 & 58.19 & 70.65 & 53.20 & 71.37 \\
& \small{\texttt{OPT-1.3B}} & 73.27 & 66.76 & 70.56 & 61.37 & 71.64 & 54.51 & 72.79 \\
& \small{\texttt{Falcon-7B}} & 77.85 & 68.01 & 73.84 & 63.70 & 75.43 & 56.75 & 77.00 \\
& \small{\texttt{Qwen2-7B}} & 81.68 & 71.68 & 75.23 & 63.20 & 79.00 & 56.41 & 80.81 \\
\midrule

\multirow{4}{*}{\textbf{DetectGPT}} 
& \small{\texttt{GPT2-XL}} & 72.51 & 68.19 & 71.51 & 70.82 & 71.81 & 68.17 & 71.98 \\
& \small{\texttt{OPT-1.3B}} & 74.50 & 70.40 & 73.04 & 77.60 & 72.83 & 72.93 & 73.30 \\
& \small{\texttt{Falcon-7B}} & 76.74 & 69.53 & 74.95 & 77.85 & 74.29 & 74.60 & 74.68 \\
& \small{\texttt{Qwen2-7B}} & 78.34 & 73.19 & 75.48 & 78.25 & 75.29 & 75.53 & 75.62 \\
\midrule 

\textbf{RADAR} & & 64.13 & 55.84 & 60.73 & 22.19 & 81.38 & 27.00 & 78.27 \\
\textbf{MAGE} & & 78.66 & 59.90 & 71.01 & 67.36 & 64.61 & 72.90 & 59.47 \\
\textbf{AIGC MPU} & & 75.30 & 71.70 & 55.86 & 56.28 & 72.17 & 47.66 & 77.83 \\
\textbf{OpenAI RoBERTa-base} & & 74.60 & 71.41 & 52.85 & 71.32 & 56.89 & 70.81 & 58.30 \\
\textbf{OpenAI RoBERTa-large} & & 70.45 & 69.87 & 51.16 & 63.07 & 60.52 & 56.74 & 66.00 \\
\bottomrule
\end{tabular}
}
\caption{The AUROC scores (\%) of the 33 MGT detectors' configurations on Beemo in the \textbf{Generation} category. A random classifier has an AUROC of 50\%. \textbf{H}=human-written; \textbf{M}=machine-generated; \textbf{E}=expert-edited; \textbf{L}=\llamaxl-edited; \textbf{G}=\gpt-edited. \textbf{L} and \textbf{G} are aggregated over three prompts (\S\ref{subsec:llm_editing}).}
\label{tab:results_generation}
\end{table*}

\begin{table*}[ht!]
\centering
\setlength{\tabcolsep}{10pt}
{\renewcommand{\arraystretch}{1}}
\resizebox{\textwidth}{!}{
\begin{tabular}{llRRRRRRR}
\toprule
\multicolumn{2}{p{2ex}}{\textbf{Detector}} & \multicolumn{1}{c}{\textbf{H vs. M}} & \multicolumn{1}{c}{\textbf{E vs. M}} & \multicolumn{1}{c}{\textbf{H vs. E}} & \multicolumn{1}{c}{\textbf{L vs. M}} & \multicolumn{1}{c}{\textbf{H vs. L}} & \multicolumn{1}{c}{\textbf{G vs. M}} & \multicolumn{1}{c}{\textbf{H vs. G}} \\
\midrule

\multicolumn{9}{c}{\textbf{Zero-shot MGT Detectors}} \\ \midrule

\textbf{Binoculars} &  & 85.97 & 81.31 & 56.46 & 74.09 & 69.67 & 70.06 & 69.48 \\ \midrule

\multirow{4}{*}{\textbf{Log Probability}} 
 & \small{\texttt{GPT2-XL}}& 76.15 & 75.18 & 50.95 & 70.37 & 60.05 & 60.10 & 70.79 \\
 & \small{\texttt{OPT-1.3B}} & 79.22 & 76.83 & 53.02 & 73.49 & 61.33 & 62.16 & 73.14 \\
 & \small{\texttt{Falcon-7B}} & 92.30 & 75.66 & 80.35 & 72.94 & 86.18 & 58.85 & 92.81 \\
 & \small{\texttt{Qwen2-7B}}  & 93.68 & 77.76 & 83.45 & 59.25 & 90.27 & 68.51 & 86.86 \\
\midrule

\multirow{4}{*}{\textbf{Rank}} 
& \small{\texttt{GPT2-XL}}& 63.60 & 62.21 & 51.78 & 62.24 & 51.17 & 56.24 & 57.96 \\
& \small{\texttt{OPT-1.3B}} & 66.54 & 64.21 & 53.64 & 65.54 & 51.78 & 56.92 & 61.35 \\
& \small{\texttt{Falcon-7B}} & 77.41 & 62.01 & 68.83 & 65.28 & 64.02 & 54.43 & 75.68 \\
& \small{\texttt{Qwen2-7B}}  & 77.53 & 60.33 & 72.00 & 53.44 & 75.69 & 56.89 & 73.84 \\
\midrule

\multirow{4}{*}{\textbf{Log-Rank}} 
& \small{\texttt{GPT2-XL}}& 77.45 & 76.13 & 51.31 & 72.50 & 59.18 & 61.92 & 70.68 \\
& \small{\texttt{OPT-1.3B}} & 78.12 & 76.84 & 51.38 & 74.30 & 58.45 & 63.33 & 70.66 \\
& \small{\texttt{Falcon-7B}} & 91.79 & 75.65 & 79.41 & 73.96 & 84.35 & 60.07 & 91.81 \\
& \small{\texttt{Qwen2-7B}}  & 93.30 & 77.78 & 82.40 & 59.26 & 89.67 & 68.52 & 86.03 \\
\midrule

\multirow{4}{*}{\textbf{Entropy}} 
& \small{\texttt{GPT2-XL}}& 35.06 & 34.36 & 50.88 & 35.35 & 48.77 & 44.33 & 39.67 \\
& \small{\texttt{OPT-1.3B}} & 37.08 & 35.25 & 51.85 & 34.43 & 51.53 & 42.63 & 43.20 \\
& \small{\texttt{Falcon-7B}} & 12.26 & 32.29 & 20.55 & 31.01 & 19.34 & 40.37 & 13.64 \\
& \small{\texttt{Qwen2-7B}}  & 6.10 & 24.96 & 12.57 & 41.65 & 8.26 & 33.31 & 10.41 \\
\midrule

\multirow{4}{*}{\makecell[l]{\textbf{DetectLLM} \\ \textbf{Likelihood Log-Rank Ratio}}} 
& \small{\texttt{GPT2-XL}}& 74.80 & 73.75 & 50.88 & 72.67 & 54.49 & 62.71 & 65.20 \\
& \small{\texttt{OPT-1.3B}} & 69.69 & 72.35 & 46.67 & 72.13 & 48.71 & 62.79 & 59.14 \\
& \small{\texttt{Falcon-7B}} & 84.28 & 71.15 & 69.35 & 72.70 & 70.31 & 61.47 & 79.23 \\
& \small{\texttt{Qwen2-7B}}  & 86.58 & 72.14 & 71.58 & 57.38 & 81.58 & 64.76 & 76.58 \\
\midrule

\multirow{4}{*}{\makecell[l]{\textbf{DetectLLM} \\ \textbf{Normalized Perturbed Log-Rank}}} 
& \small{\texttt{GPT2-XL}}& 55.89 & 67.17 & 53.56 & 59.94 & 54.54 & 50.94 & 55.97 \\
& \small{\texttt{OPT-1.3B}} & 56.81 & 68.31 & 54.02 & 64.00 & 54.75 & 52.58 & 56.60 \\
& \small{\texttt{Falcon-7B}} & 63.13 & 68.05 & 58.78 & 64.72 & 59.87 & 52.75 & 62.89 \\
& \small{\texttt{Qwen2-7B}}  & 68.05 & 72.53 & 60.61 & 64.39 & 64.04 & 51.52 & 67.94 \\
\midrule

\multirow{4}{*}{\textbf{DetectGPT}} 
& \small{\texttt{GPT2-XL}}& 58.80 & 69.55 & 55.31 & 62.10 & 56.40 & 58.44 & 57.30 \\
& \small{\texttt{OPT-1.3B}} & 61.25 & 75.15 & 56.14 & 69.21 & 57.05 & 63.20 & 58.59 \\
& \small{\texttt{Falcon-7B}} & 65.51 & 71.80 & 60.04 & 66.81 & 60.93 & 62.49 & 62.44 \\
& \small{\texttt{Qwen2-7B}}  & 68.14 & 74.24 & 61.11 & 62.56 & 64.31 & 62.64 & 64.70 \\
\midrule

\textbf{RADAR} & &  47.78 & 46.69 & 51.43 & 37.84 & 57.84 & 40.62 & 55.58 \\
\textbf{MAGE} & &  76.05 & 64.39 & 63.31 & 58.99 & 70.21 & 62.23 & 67.34 \\
\textbf{AIGC MPU} & &  75.33 & 79.87 & 40.69 & 85.27 & 31.55 & 80.63 & 39.59 \\
\textbf{OpenAI RoBERTa-base} & &  76.68 & 76.55 & 48.95 & 80.72 & 41.85 & 78.99 & 43.93 \\
\textbf{OpenAI RoBERTa-large} & &  72.60 & 71.87 & 51.46 & 78.94 & 40.87 & 73.95 & 47.36 \\
\bottomrule
\end{tabular}
}
\caption{The AUROC scores (\%) of the 33 MGT detectors' configurations on Beemo in the \textbf{Open QA} category. A random classifier has an AUROC of 50\%. \textbf{H}=human-written; \textbf{M}=machine-generated; \textbf{E}=expert-edited; \textbf{L}=\llamaxl-edited; \textbf{G}=\gpt-edited. \textbf{L} and \textbf{G} are aggregated over three prompts (\S\ref{subsec:llm_editing}).}
\label{tab:results_openqa}
\end{table*}

\begin{table*}[ht!]
\centering
\setlength{\tabcolsep}{10pt}
{\renewcommand{\arraystretch}{1}}
\resizebox{\textwidth}{!}{
\begin{tabular}{llRRRRRRR}
\toprule
\multicolumn{2}{p{2ex}}{\textbf{Detector}} & \multicolumn{1}{c}{\textbf{H vs. M}} & \multicolumn{1}{c}{\textbf{E vs. M}} & \multicolumn{1}{c}{\textbf{H vs. E}} & \multicolumn{1}{c}{\textbf{L vs. M}} & \multicolumn{1}{c}{\textbf{H vs. L}} & \multicolumn{1}{c}{\textbf{G vs. M}} & \multicolumn{1}{c}{\textbf{H vs. G}} \\
\midrule

\multicolumn{9}{c}{\textbf{Zero-shot MGT Detectors}} \\ \midrule

\textbf{Binoculars} &  & 80.25 & 76.05 & 55.97 & 38.28 & 88.65 & 39.19 & 85.97 \\ 
\midrule

\multirow{4}{*}{\textbf{Log Probability}} 
 & \small{\texttt{GPT2-XL}}& 70.21 & 64.29 & 57.11 & 54.74 & 67.61 & 59.37 & 61.70 \\
 & \small{\texttt{OPT-1.3B}} & 74.95 & 66.61 & 60.53 & 57.90 & 70.88 & 59.25 & 67.11 \\
 & \small{\texttt{Falcon-7B}} & 87.88 & 63.56 & 81.00 & 54.90 & 88.29 & 55.22 & 84.99 \\
 & \small{\texttt{Qwen2-7B}}  & 90.41 & 67.70 & 81.50 & 55.90 & 87.44 & 61.80 & 84.47 \\
\midrule

\multirow{4}{*}{\textbf{Rank}}  
& \small{\texttt{GPT2-XL}}& 57.24 & 51.93 & 55.43 & 53.79 & 54.16 & 53.28 & 54.16 \\
& \small{\texttt{OPT-1.3B}} & 62.02 & 53.70 & 58.76 & 57.24 & 55.81 & 52.42 & 59.73 \\
& \small{\texttt{Falcon-7B}} & 70.81 & 49.83 & 70.69 & 55.17 & 66.15 & 49.63 & 70.25 \\
& \small{\texttt{Qwen2-7B}}  & 70.71 & 51.04 & 70.21 & 50.35 & 70.54 & 50.69 & 70.38 \\
\midrule 

\multirow{4}{*}{\textbf{Log-Rank}} 
& \small{\texttt{GPT2-XL}}& 69.88 & 63.95 & 57.05 & 56.47 & 65.67 & 60.01 & 60.51 \\
& \small{\texttt{OPT-1.3B}} & 73.26 & 65.56 & 59.45 & 58.76 & 67.94 & 59.99 & 64.40 \\
& \small{\texttt{Falcon-7B}} & 87.02 & 62.35 & 80.41 & 56.08 & 86.54 & 56.16 & 83.39 \\
& \small{\texttt{Qwen2-7B}}  & 90.04 & 66.86 & 81.06 & 55.62 & 87.05 & 61.24 & 84.05 \\
\midrule

\multirow{4}{*}{\textbf{Entropy}} 
& \small{\texttt{GPT2-XL}}& 35.56 & 42.34 & 42.83 & 42.84 & 41.56 & 44.49 & 40.63 \\
& \small{\texttt{OPT-1.3B}} & 34.05 & 42.23 & 41.07 & 41.60 & 40.85 & 44.02 & 39.29 \\
& \small{\texttt{Falcon-7B}} & 16.02 & 45.02 & 18.24 & 45.45 & 15.79 & 47.90 & 16.29 \\
& \small{\texttt{Qwen2-7B}}  & 5.78 & 29.58 & 13.81 & 43.19 & 8.46 & 36.39 & 11.13 \\
\midrule 

\multirow{4}{*}{\makecell[l]{\textbf{DetectLLM} \\ \textbf{Likelihood Log-Rank Ratio}}} 
& \small{\texttt{GPT2-XL}}& 64.75 & 59.97 & 55.57 & 59.14 & 56.95 & 59.89 & 54.74 \\
& \small{\texttt{OPT-1.3B}} & 64.29 & 60.76 & 54.67 & 60.52 & 55.65 & 60.95 & 53.84 \\
& \small{\texttt{Falcon-7B}} & 78.55 & 56.93 & 73.60 & 58.83 & 73.77 & 58.20 & 71.78 \\
& \small{\texttt{Qwen2-7B}}  & 83.14 & 62.02 & 74.66 & 54.01 & 80.31 & 58.01 & 77.49 \\
\midrule 

\multirow{4}{*}{\makecell[l]{\textbf{DetectLLM} \\ \textbf{Normalized Perturbed Log-Rank}}} 
& \small{\texttt{GPT2-XL}}& 61.93 & 56.39 & 60.78 & 45.90 & 62.60 & 49.25 & 62.08 \\
& \small{\texttt{OPT-1.3B}} & 63.48 & 59.90 & 61.74 & 51.23 & 63.23 & 49.81 & 63.60 \\
& \small{\texttt{Falcon-7B}} & 68.11 & 58.33 & 66.39 & 49.17 & 68.32 & 47.17 & 68.82 \\
& \small{\texttt{Qwen2-7B}}  & 72.17 & 66.17 & 68.07 & 45.76 & 72.91 & 47.08 & 72.89 \\
\midrule 

\multirow{4}{*}{\textbf{DetectGPT}} 
& \small{\texttt{GPT2-XL}}& 60.80 & 59.71 & 59.88 & 46.92 & 62.09 & 53.66 & 60.98 \\
& \small{\texttt{OPT-1.3B}} & 62.98 & 66.23 & 60.66 & 52.93 & 62.99 & 55.29 & 62.72 \\
& \small{\texttt{Falcon-7B}} & 67.54 & 67.08 & 64.56 & 53.33 & 67.41 & 56.32 & 66.91 \\
& \small{\texttt{Qwen2-7B}}  & 70.40 & 72.84 & 65.65 & 48.42 & 71.07 & 56.65 & 69.44 \\
\midrule

\textbf{RADAR} & &  41.12 & 46.81 & 44.16 & 41.96 & 46.88 & 45.31 & 44.75 \\
\textbf{MAGE} & &  69.88 & 59.94 & 61.09 & 46.81 & 72.77 & 48.72 & 70.96 \\
\textbf{AIGC MPU} & &  71.85 & 68.09 & 54.00 & 70.64 & 50.94 & 64.91 & 56.09 \\
\textbf{OpenAI RoBERTa-base} & &  49.03 & 55.56 & 42.30 & 55.24 & 42.88 & 55.23 & 43.43 \\
\textbf{OpenAI RoBERTa-large} & &  47.66 & 52.87 & 44.33 & 55.00 & 42.13 & 52.18 & 45.54 \\
\bottomrule
\end{tabular}
}
\caption{The AUROC scores (\%) of the 33 MGT detectors' configurations on Beemo in the \textbf{Summarize} category. A random classifier has an AUROC of 50\%. \textbf{H}=human-written; \textbf{M}=machine-generated; \textbf{E}=expert-edited; \textbf{L}=\llamaxl-edited; \textbf{G}=\gpt-edited. \textbf{L} and \textbf{G} are aggregated over three prompts (\S\ref{subsec:llm_editing}).}
\label{tab:results_summarize}
\end{table*}

\begin{table*}[ht!]
\centering
\setlength{\tabcolsep}{10pt}
{\renewcommand{\arraystretch}{1}}
\resizebox{\textwidth}{!}{
\begin{tabular}{llRRRRRRR}
\toprule
\multicolumn{2}{p{2ex}}{\textbf{Detector}} & \multicolumn{1}{c}{\textbf{H vs. M}} & \multicolumn{1}{c}{\textbf{E vs. M}} & \multicolumn{1}{c}{\textbf{H vs. E}} & \multicolumn{1}{c}{\textbf{L vs. M}} & \multicolumn{1}{c}{\textbf{H vs. L}} & \multicolumn{1}{c}{\textbf{G vs. M}} & \multicolumn{1}{c}{\textbf{H vs. G}} \\
\midrule

\multicolumn{9}{c}{\textbf{Zero-shot MGT Detectors}} \\ \midrule

\textbf{Binoculars} & &  83.83 & 77.66 & 61.90 & 49.35 & 82.51 & 46.54 & 81.91 \\
\midrule

\multirow{4}{*}{\textbf{Log Probability}} 
 & \small{\texttt{GPT2-XL}}& 66.56 & 61.63 & 56.03 & 63.39 & 54.23 & 62.44 & 54.77 \\
 & \small{\texttt{OPT-1.3B}} & 72.73 & 62.53 & 62.20 & 65.91 & 59.42 & 62.31 & 62.24 \\
 & \small{\texttt{Falcon-7B}} & 84.17 & 62.26 & 77.70 & 64.38 & 77.68 & 61.58 & 78.08 \\
 & \small{\texttt{Qwen2-7B}}  & 84.78 & 65.91 & 75.18 & 55.30 & 81.58 & 60.61 & 78.38 \\
\midrule 

\multirow{4}{*}{\textbf{Rank}} 
& \small{\texttt{GPT2-XL}}& 56.44 & 52.87 & 54.43 & 57.68 & 48.82 & 56.91 & 49.23 \\
& \small{\texttt{OPT-1.3B}} & 64.11 & 53.05 & 62.36 & 61.82 & 53.20 & 56.43 & 58.19 \\
& \small{\texttt{Falcon-7B}} & 72.94 & 53.61 & 71.12 & 61.98 & 62.50 & 56.47 & 67.63 \\
& \small{\texttt{Qwen2-7B}}  & 73.23 & 56.84 & 68.20 & 52.28 & 71.55 & 54.56 & 69.88 \\
\midrule

\multirow{4}{*}{\textbf{Log-Rank}} 
& \small{\texttt{GPT2-XL}}& 66.19 & 61.05 & 56.18 & 64.29 & 52.51 & 62.85 & 53.66 \\
& \small{\texttt{OPT-1.3B}} & 70.04 & 61.91 & 59.80 & 66.24 & 55.44 & 62.98 & 58.26 \\
& \small{\texttt{Falcon-7B}} & 82.13 & 61.80 & 75.54 & 65.03 & 73.98 & 62.24 & 74.56 \\
& \small{\texttt{Qwen2-7B}}  & 83.28 & 65.37 & 73.57 & 55.12 & 80.04 & 60.25 & 76.81 \\
\midrule

\multirow{4}{*}{\textbf{Entropy}} 
& \small{\texttt{GPT2-XL}}& 45.41 & 47.10 & 48.15 & 36.49 & 58.94 & 41.54 & 53.82 \\
& \small{\texttt{OPT-1.3B}} & 42.17 & 47.53 & 44.18 & 34.54 & 57.56 & 40.72 & 51.19 \\
& \small{\texttt{Falcon-7B}} & 26.84 & 46.09 & 28.43 & 36.53 & 37.46 & 41.45 & 33.24 \\
& \small{\texttt{Qwen2-7B}}  & 14.62 & 34.95 & 22.56 & 44.98 & 17.27 & 39.97 & 19.91 \\
\midrule

\multirow{4}{*}{\makecell[l]{\textbf{DetectLLM} \\ \textbf{Likelihood Log-Rank Ratio}}} 
& \small{\texttt{GPT2-XL}}& 63.75 & 58.94 & 55.98 & 65.69 & 47.73 & 63.35 & 49.88 \\
& \small{\texttt{OPT-1.3B}} & 58.88 & 58.20 & 51.28 & 65.42 & 43.10 & 63.18 & 45.27 \\
& \small{\texttt{Falcon-7B}} & 71.86 & 58.70 & 65.68 & 65.17 & 58.81 & 63.02 & 60.02 \\
& \small{\texttt{Qwen2-7B}}  & 74.21 & 61.41 & 65.15 & 53.80 & 71.19 & 57.61 & 68.17 \\
\midrule 

\multirow{4}{*}{\makecell[l]{\textbf{DetectLLM} \\ \textbf{Normalized Perturbed Log-Rank}}} 
& \small{\texttt{GPT2-XL}}& 64.30 & 62.54 & 62.24 & 50.66 & 64.13 & 50.25 & 64.31 \\
& \small{\texttt{OPT-1.3B}} & 65.81 & 63.51 & 63.35 & 52.73 & 65.25 & 50.47 & 65.87 \\
& \small{\texttt{Falcon-7B}} & 70.27 & 63.41 & 66.98 & 53.06 & 69.74 & 51.08 & 70.34 \\
& \small{\texttt{Qwen2-7B}}  & 73.57 & 69.32 & 67.83 & 48.16 & 74.28 & 49.63 & 74.00 \\
\midrule

\multirow{4}{*}{\textbf{DetectGPT}} 
& \small{\texttt{GPT2-XL}}& 64.00 & 63.18 & 61.76 & 60.89 & 62.23 & 61.87 & 61.97 \\
& \small{\texttt{OPT-1.3B}} & 66.43 & 66.27 & 63.51 & 66.98 & 63.61 & 65.36 & 63.79 \\
& \small{\texttt{Falcon-7B}} & 69.91 & 64.56 & 67.07 & 69.00 & 66.68 & 68.00 & 66.68 \\
& \small{\texttt{Qwen2-7B}}  & 71.48 & 69.51 & 67.43 & 64.94 & 68.96 & 67.85 & 68.12 \\
\midrule

\textbf{RADAR} & &  57.24 & 54.46 & 53.01 & 24.69 & 77.32 & 27.91 & 74.72 \\
\textbf{MAGE} & &  70.67 & 59.56 & 62.05 & 59.23 & 62.77 & 60.29 & 61.34 \\
\textbf{AIGC MPU} & &  71.29 & 69.80 & 52.30 & 52.49 & 70.62 & 47.35 & 73.71 \\
\textbf{OpenAI RoBERTa-base} & &  65.61 & 64.11 & 51.08 & 58.16 & 59.77 & 58.70 & 58.83 \\
\textbf{OpenAI RoBERTa-large }& &  63.01 & 62.77 & 49.84 & 56.49 & 57.51 & 52.17 & 61.49 \\
\bottomrule
\end{tabular}
}
\caption{The AUROC scores (\%) of the 33 MGT detectors' configurations on Beemo in the \textbf{Rewrite} category. A random classifier has an AUROC of 50\%. \textbf{H}=human-written; \textbf{M}=machine-generated; \textbf{E}=expert-edited; \textbf{L}=\llamaxl-edited; \textbf{G}=\gpt-edited. \textbf{L} and \textbf{G} are aggregated over three prompts (\S\ref{subsec:llm_editing}).}
\label{tab:results_rewrite}
\end{table*}

\begin{table*}[ht!]
\centering
\setlength{\tabcolsep}{10pt}
{\renewcommand{\arraystretch}{1}}
\resizebox{\textwidth}{!}{
\begin{tabular}{llRRRRRRR}
\toprule
\multicolumn{2}{p{2ex}}{\textbf{Detector}} & \multicolumn{1}{c}{\textbf{H vs. M}} & \multicolumn{1}{c}{\textbf{E vs. M}} & \multicolumn{1}{c}{\textbf{H vs. E}} & \multicolumn{1}{c}{\textbf{L vs. M}} & \multicolumn{1}{c}{\textbf{H vs. L}} & \multicolumn{1}{c}{\textbf{G vs. M}} & \multicolumn{1}{c}{\textbf{H vs. G}} \\
\midrule

\multicolumn{9}{c}{\textbf{Zero-shot MGT Detectors}} \\ \midrule

\textbf{Binoculars} & &  74.83 & 76.09 & 49.49 & 52.07 & 75.44 & 47.29 & 76.63 \\
\midrule

\multirow{4}{*}{\textbf{Log Probability}} 
 & \small{\texttt{GPT2-XL}}& 70.06 & 63.69 & 57.60 & 54.76 & 68.95 & 54.42 & 67.08 \\
 & \small{\texttt{OPT-1.3B}} & 71.61 & 65.11 & 58.35 & 56.31 & 69.54 & 54.34 & 68.67 \\
 & \small{\texttt{Falcon-7B}} & 83.29 & 60.60 & 76.37 & 53.72 & 83.81 & 51.56 & 82.89 \\
 & \small{\texttt{Qwen2-7B}}  & 87.08 & 64.19 & 76.77 & 54.73 & 83.64 & 59.46 & 80.21 \\
\midrule 

\multirow{4}{*}{\textbf{Rank}} 
& \small{\texttt{GPT2-XL}}& 54.86 & 54.46 & 50.74 & 52.09 & 53.42 & 52.24 & 52.94 \\
& \small{\texttt{OPT-1.3B}} & 58.22 & 57.86 & 51.15 & 55.57 & 54.44 & 53.75 & 55.08 \\
& \small{\texttt{Falcon-7B}} & 64.51 & 55.45 & 59.75 & 54.25 & 61.84 & 50.23 & 64.21 \\
& \small{\texttt{Qwen2-7B}}  & 65.20 & 55.52 & 59.51 & 51.84 & 63.30 & 53.68 & 61.41 \\
\midrule 

\multirow{4}{*}{\textbf{Log-Rank}} 
& \small{\texttt{GPT2-XL}}& 70.84 & 63.71 & 58.69 & 56.00 & 68.61 & 54.63 & 67.50 \\
& \small{\texttt{OPT-1.3B}} & 70.90 & 64.72 & 57.83 & 56.64 & 68.26 & 54.80 & 67.49 \\
& \small{\texttt{Falcon-7B}} & 82.52 & 60.46 & 75.92 & 54.05 & 82.97 & 52.14 & 81.66 \\
& \small{\texttt{Qwen2-7B}}  & 87.11 & 64.87 & 76.75 & 54.96 & 83.66 & 59.91 & 80.20 \\
\midrule 

\multirow{4}{*}{\textbf{Entropy}} 
& \small{\texttt{GPT2-XL}}& 31.34 & 42.60 & 37.21 & 46.67 & 31.46 & 48.52 & 31.83 \\
& \small{\texttt{OPT-1.3B}} & 31.09 & 41.61 & 38.04 & 46.15 & 31.67 & 48.47 & 31.60 \\
& \small{\texttt{Falcon-7B}} & 18.33 & 46.14 & 19.95 & 49.14 & 16.26 & 50.68 & 16.97 \\
& \small{\texttt{Qwen2-7B}}  & 10.18 & 38.04 & 15.74 & 46.01 & 12.03 & 42.03 & 13.89 \\
\midrule 

\multirow{4}{*}{\makecell[l]{\textbf{DetectLLM} \\ \textbf{Likelihood Log-Rank Ratio}}} 
& \small{\texttt{GPT2-XL}}& 69.18 & 60.24 & 59.98 & 57.93 & 64.22 & 54.47 & 65.69 \\
& \small{\texttt{OPT-1.3B}} & 65.08 & 61.43 & 55.58 & 56.15 & 61.35 & 54.87 & 61.09 \\
& \small{\texttt{Falcon-7B}} & 74.48 & 58.24 & 68.55 & 53.86 & 73.54 & 53.23 & 72.05 \\
& \small{\texttt{Qwen2-7B}}  & 80.44 & 62.98 & 70.09 & 54.33 & 76.99 & 58.65 & 73.54 \\
\midrule 

\multirow{4}{*}{\makecell[l]{\textbf{DetectLLM} \\ \textbf{Normalized Perturbed Log-Rank}}} 
& \small{\texttt{GPT2-XL}}& 62.14 & 62.06 & 60.72 & 56.56 & 61.28 & 55.24 & 61.61 \\
& \small{\texttt{OPT-1.3B}} & 62.62 & 61.25 & 61.30 & 56.50 & 61.69 & 52.51 & 62.34 \\
& \small{\texttt{Falcon-7B}} & 64.85 & 58.15 & 63.51 & 52.37 & 64.32 & 52.22 & 64.59 \\
& \small{\texttt{Qwen2-7B}}  & 68.00 & 66.66 & 64.54 & 52.47 & 67.06 & 55.54 & 67.17 \\
\midrule 

\multirow{4}{*}{\textbf{DetectGPT}} 
& \small{\texttt{GPT2-XL}}& 63.87 & 60.33 & 61.14 & 52.12 & 62.50 & 53.01 & 62.68 \\
& \small{\texttt{OPT-1.3B}} & 65.05 & 63.49 & 62.05 & 55.13 & 63.12 & 52.66 & 63.94 \\
& \small{\texttt{Falcon-7B}} & 67.20 & 60.53 & 64.05 & 50.31 & 65.62 & 52.63 & 65.71 \\
& \small{\texttt{Qwen2-7B}}  & 68.80 & 66.55 & 64.67 & 46.95 & 67.81 & 55.01 & 66.92 \\
\midrule 

\textbf{RADAR} & &  39.41 & 52.57 & 34.99 & 71.24 & 13.34 & 74.10 & 11.02 \\
\textbf{MAGE} & &  71.88 & 64.21 & 58.50 & 46.43 & 74.24 & 43.57 & 77.66 \\
\textbf{AIGC MPU} & &  76.62 & 73.45 & 52.60 & 68.83 & 56.73 & 65.70 & 58.58 \\
\textbf{OpenAI RoBERTa-base} & &  57.17 & 62.46 & 42.64 & 65.45 & 38.65 & 70.30 & 33.11 \\
\textbf{OpenAI RoBERTa-large} & &  48.72 & 57.38 & 40.69 & 74.68 & 22.41 & 69.15 & 28.56 \\
\bottomrule
\end{tabular}
}
\caption{The AUROC scores (\%) of the 33 MGT detectors' configurations on Beemo in the \textbf{Closed QA} category. A random classifier has an AUROC of 50\%. \textbf{H}=human-written; \textbf{M}=machine-generated; \textbf{E}=expert-edited; \textbf{L}=\llamaxl-edited; \textbf{G}=\gpt-edited. \textbf{L} and \textbf{G} are aggregated over three prompts (\S\ref{subsec:llm_editing}).}
\label{tab:results_closedqa}
\end{table*}